\documentclass[10pt, a4paper, logo]{googledeepmind} %
\usepackage{times}

\usepackage{hyperref}
\usepackage[leftcaption]{sidecap} % caption 在右边

\usepackage{amsmath,amsfonts,bm}
\usepackage{natbib}
\usepackage{fancyhdr}
\def\eqref#1{equation~\ref{#1}}
\def\1{\bm{1}}

\DeclareMathAlphabet{\mathsfit}{\encodingdefault}{\sfdefault}{m}{sl}
\SetMathAlphabet{\mathsfit}{bold}{\encodingdefault}{\sfdefault}{bx}{n}

\usepackage{hyperref}
\usepackage{url}
\usepackage{booktabs}
\usepackage{graphicx}
\usepackage{subcaption}
\usepackage{amssymb}
\usepackage[most]{tcolorbox}
\usepackage[table]{xcolor}
\tcbuselibrary{breakable} % 加载分页支持库
\usepackage{amsmath}
\usepackage{siunitx}
\usepackage{colortbl}

\usepackage{microtype}
\usepackage{color}
\usepackage{wrapfig}
\usepackage{natbib}
\usepackage{colortbl}
\usepackage{microtype}
\usepackage{graphicx}
\usepackage{booktabs} % for professional tables
\usepackage{array}
\usepackage{textcomp}
\usepackage{stfloats}
\usepackage{float}
\usepackage{verbatim}
\usepackage[ruled, linesnumbered, lined]{algorithm2e}
\usepackage{multirow}
\usepackage{enumitem}

\definecolor{BestColor}{HTML}{C8E6C9}  % 一个柔和的绿色
\definecolor{SecondBestColor}{HTML}{FFF9C4} % 一个非常淡的黄色

\definecolor{plancolor}{HTML}{2874A6}
\definecolor{factcolor}{HTML}{2E7D32}
\definecolor{answercolor}{HTML}{B71C1C}
\definecolor{findingcolor}{HTML}{B7791F}
\definecolor{boxbg}{HTML}{FDF6EC}
\definecolor{boxframe}{HTML}{D4A574}

\definecolor{red3}{HTML}{C52A20}
\definecolor{red2}{HTML}{B36A6F}
\definecolor{red1}{HTML}{FFb5b5}
\definecolor{purple}{HTML}{B36A6F}
\definecolor{purple1}{HTML}{8d3a94}
\definecolor{purplex}{HTML}{9564bf}
\definecolor{darkyellow}{HTML}{D5BA82}
\definecolor{blue1}{HTML}{508AB2}
\definecolor{blue2}{HTML}{C4E4E3}
\definecolor{green1}{HTML}{A1D0C7}
\definecolor{green2}{HTML}{BFF6BA}
\definecolor{green3}{HTML}{028100}
\definecolor{teal}{HTML}{508AB2}

\newtcbtheorem[]{protocolexmp}{Example}%
{colback=purplex!5,colframe=purplex!80,fonttitle=\bfseries, left=.08in, right=.08in, bottom=.05in, top=.05in}{protocolexmp}

\usepackage{tcolorbox}
\usepackage{amsmath,amsfonts}

\definecolor{ggg}{RGB}{26,179,0}
\definecolor{rrr}{RGB}{179,0,0}
\definecolor{oodc}{RGB}{31,73,121}
\definecolor{idc}{RGB}{68,142,68}

\definecolor{mygray}{gray}{0.9}

\def\Bias#1#2{\bm{b}}
\newtcolorbox{examplebox}[2][]{ % 允许传入可选参数 [#1] 和必选标题参数 {#2}
    breakable, % 关键：允许跨页分割
    enhanced, % 增强模式（可选，支持更多样式）
    colback=white, % 框体内背景色
    colframe=cyan, % 边框颜色
    coltitle=white, % 标题文字颜色
    fonttitle=\bfseries, % 标题字体加粗
    title=#2, % 框体标题（第二个必选参数）
    overlay middle={\draw[cyan, line width=1pt](frame.south west)--(frame.south east);}, % 分割处添加横线
    overlay last={\draw[cyan, line width=1pt](frame.south west)--(frame.south east);}, % 最后一页底部横线
    #1 % 允许在调用时传入其他可选参数以覆盖默认样式
}

\usepackage[T1]{fontenc}
\usepackage{booktabs}      % 用于漂亮的表格线 (toprule, midrule, etc.)
\usepackage{graphicx}      % 用于 \resizebox
\usepackage[table]{xcolor} % 用于颜色
\usepackage{siunitx}       % 用于 S 列，实现小数点对齐
\usepackage{etoolbox}      % 用于 \ifstrequal，实现条件判断
\usepackage[normalem]{ulem}     % 用于更灵活的下划线 \uline

\definecolor{impcolor}{HTML}{2E8B57} % 提升使用的海绿色 (SeaGreen)

\newcommand{\improvementstyle}[1]{$^{\textcolor{impcolor}{\tiny #1}}$}

\newcommand{\scoreimp}[2]{%
  \textbf{#1}%
  \ifstrequal{#2}{+0.0}{}{%
    \ifstrequal{#2}{0.0}{}{%
      \makebox[0pt][l]{\improvementstyle{#2}}%
    }%
  }%
}

\title{From Proprietary to Open-Source: Bridging the Distribution Gap via Multi-Agent Protocol Distillation in Agentic Search}

\author[1*\dag]{Junlin Liu}
\author[2*]{Jiangwang Chen}
\author[2*]{Zixin Song}
\author[3*]{Shuaiyu Zhou}
\author[4]{Chunji Lv}
\author[5]{Hank Wu}
\author[6]{\protect\linebreak[4]\mbox{Kailin Jiang}}
\author[2]{\mbox{Jinyang Wu}}
\author[1]{Bohan Yu}
\author[1]{Chenxi Zhou}

\affil[1]{University of Chinese Academy of Sciences}
\affil[2]{Tsinghua University}
\affil[3]{Peking University}
\affil[4]{Beijing Institute of Technology}
\affil[5]{\mbox{East China Normal University}}
\affil[6]{University of Science and Technology of China}

\begin{abstract}
Agentic search enables large language models to solve knowledge-intensive tasks by interleaving multi-step reasoning with retrieval, yet optimizing this with outcome-based reinforcement learning (RL) provides only sparse supervision. Knowledge distillation can supply denser guidance, and advanced proprietary models with their strong reasoning capabilities are promising teachers. 
While distilling from proprietary models can densify this supervisory signal, conventional logit-matching is precluded by hidden logits and mismatched tokenizers, whereas raw natural language trajectory imitation transfers superficial stylistic artifacts rather than core reasoning competence. 
To address the heterogeneous distillation problem and bridge the distribution gap, we propose \textbf{M}ulti-\textbf{A}gent \textbf{P}rotocol \textbf{D}istillation (\textbf{MAPD}), a joint distillation and RL framework uses a structured, style-normalized protocol as an intermediate representation. An offline multi-agent system (MAS) decomposes each query, retrieves supporting evidence, repairs failed searches, and converts the resulting exploration trace into a JSON protocol containing the task type, reasoning plan, and extractive grounding facts.
During training, the protocol is provided only to a privileged branch of the student policy, whose token distributions furnish a dense distillation signal alongside the sparse RL objective. 
Extensive evaluations across seven QA benchmarks demonstrate that MAPD consistently outperforms competitive distillation and RL, achieving average success rates of $39.4\%$ on Qwen3-1.7B and $44.4\%$ on Qwen3-4B. Crucially, the framework generalizes robustly across diverse proprietary teachers while effectively mitigating the student policy from style drift and verbosity degeneration. The code is open sourced in \href{https://github.com/AaronLiu0702/MAPD}{\texttt{https://github.com/AaronLiu0702/MAPD}}
\end{abstract}

\begin{document}
\maketitle
%\vspace{-1mm}
\section{Introduction}\label{sec:introduction}
Driven by the rapid advancement of large language models (LLMs), research has shifted from single-turn reasoning to multi-turn agentic tasks, with agentic search emerging as a crucial paradigm for tackling knowledge-intensive challenges~\citep{jin2025search, zhang2025web, zheng2025deepresearcher, luo2025globalrag, liang2026search}. However, existing agentic search systems are often optimized with outcome-based reinforcement learning from verifiable rewards (RLVR), where final-answer correctness provides only sparse supervision for long interaction trajectories. Recent studies supplement RLVR with online policy distillation (OPD), using token-level distribution alignment to provide denser guidance for intermediate decisions~\citep{dong2025agentic, wang2026openclaw, lu2026self}.

Although combining RL with OPD provides a more reliable training signal, the overall efficacy of OPD strictly hinges on a high-quality teacher policy. 
To unlock the potential of open-source models in complex agentic search, leveraging state-of-the-art proprietary models as expert teachers is highly desirable, given their exceptional capabilities in dynamic task decomposition, multi-hop reasoning, and information synthesis. 
However, distilling knowledge from these proprietary teachers to open-source students presents two major bottlenecks. 
First, traditional OPD fundamentally relies on token-level KL divergence alignment. Yet, the discrepancy of heterogeneous tokenizers and the absence of accessible teacher logits render this mathematical alignment undefined~\citep{agarwal2024policy, li2026rethinking}.
Second, as a common fallback, natural language trajectory distillation frequently fails in agentic scenarios. By forcing open-source student models to mimic the verbose and idiosyncratic reasoning styles of proprietary experts, this approach exacerbates the distribution gap. It induces severe style drift and hallucinations, failing to transfer the underlying cognitive strategies while degrading performance in agentic tasks~\citep{gudibande2023false, zhao2026self}. 
Existing approaches therefore lack an intermediate representation that can convert black-box expert behavior into dense supervision without exposing the student to teacher-specific surface forms.

\begin{figure}[t]
    \centering
    \includegraphics[width=0.9\textwidth]{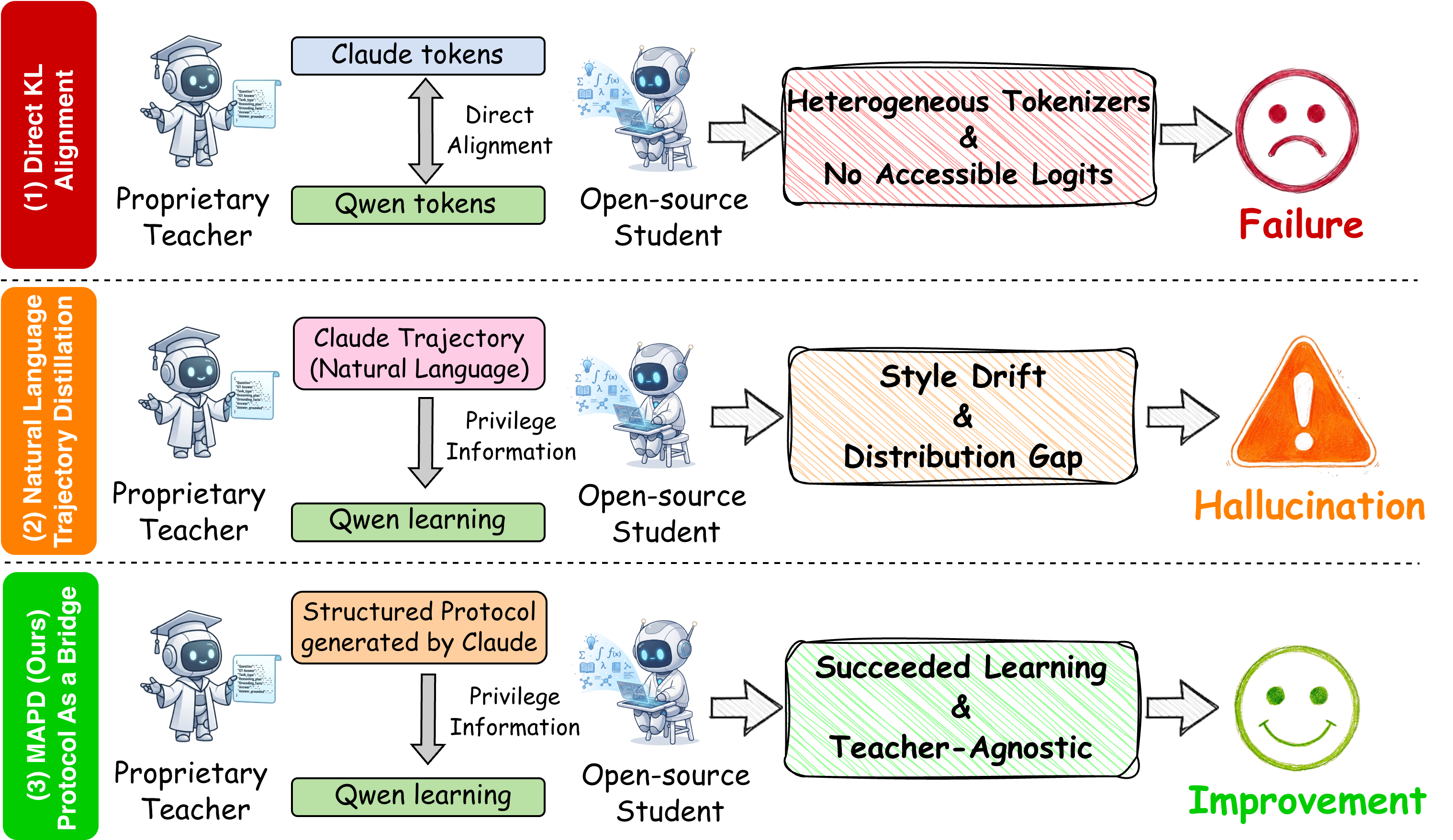} 
    \caption{Distilling from proprietary teacher to open-source student faces two bottlenecks: (1) heterogeneous tokenizers and inaccessible logits preclude direct alignment; (2) free-form natural-language trajectory imitation induces style drift and hallucination.}
    \label{fig:motivation}
\end{figure}

To address these bottlenecks, we propose \textbf{M}ulti-\textbf{A}gent \textbf{P}rotocol \textbf{D}istillation (\textbf{MAPD}), a joint distillation and RL framework built around a structured, style-normalized intermediate representation between proprietary experts and open-source students. 
MAPD separates offline protocol synthesis from online policy alignment. In the offline stage, a proprietary-model-powered MAS decomposes each query, retrieves supporting evidence, repairs failed searches, and converts the exploration trace into a JSON protocol. In the training stage, this protocol is reformatted as privileged information into a conditioned branch of the student policy. Since this privileged branch and the student branch share a same model, their distributions are aligned via a dense self-distillation objective that complements the sparse outcome-based RL objective. By distilling through structured protocols rather than free-form natural-language trajectories, MAPD reduces the student's exposure to teacher-specific verbosity and formatting artifacts. Crucially, the proprietary model and MAS are used only for offline protocol synthesis, adding zero overhead to inference. 
Our main contributions are summarized as follows:
\begin{itemize}
    \item We propose MAPD, a joint on-policy self-distillation and outcome-based RL framework designed to bridge the distribution gap between proprietary and open-source. By aligning a protocol-conditioned branch with a student branch of the same policy, it provides dense guidance without requiring access to proprietary logits or cross-tokenizer alignment.
    \item We introduce a structured, style-normalized JSON protocol as the intermediate representation to decouple core strategies from the linguistic shell. To reliably synthesize protocols, we construct a robust MAS with specialized roles and rigorous automated checks.
    \item Experiments with three top-tier proprietary models (Claude-Opus-4.6, GPT-5.5, and Gemini-3.1-Pro) as expert teachers demonstrate that MAPD yields reliable gains and these improvements can transfer across different proprietary models without retuning the pipeline.
\end{itemize}

%\vspace{-1mm}
\section{Problem Formulation \& Preliminaries}\label{sec:preliminary}

\subsection{Agentic Search}
We formulate agentic search as a multi-turn interaction paradigm. Given a question $x$, a language model policy $\pi_\theta$ interacts with a retrieval-augmented environment for up to $K$ turns. At each turn, the agent generates reasoning steps and optionally issues a retrieval query, to which the environment responds with relevant passages as the next observation. For notational simplicity, we flatten the entire generated trajectory across all turns into a single token sequence
\begin{equation}
  y = (y_1, \ldots, y_T) \sim \pi_\theta(\cdot \mid x).
\end{equation}
The interaction terminates either when the agent emits a designated final answer string or reaches the maximum turn limit $K$. Each completed episode receives a sparse, trajectory-level binary reward $R(x,y) \in \{0,1\}$, which is determined by parsing the final predicted answer from $y$ and evaluating it via strict exact-match (EM) against the ground-truth.

\subsection{Group Relative Policy Optimization (GRPO)}
For a given prompt $x$, GRPO~\citep{shao2024deepseekmath} samples a group of $G$ rollouts $\{y^{(i)}\}_{i=1}^{G}$ and computes advantages $\hat{A}^{(i)}$ derived from the terminal rewards. The policy is updated by optimizing a clipped surrogate objective $J_t^{(i)}$ alongside a per-token KL penalty against a reference model $\pi_{\text{ref}}$:
\begin{equation}
\begin{aligned}
    J_t^{(i)} &= \min \Bigl( \rho_t^{(i)} \hat{A}^{(i)},\; \text{clip}\bigl(\rho_t^{(i)}, 1-\epsilon, 1+\epsilon\bigr) \hat{A}^{(i)} \Bigr), \\
    \mathcal{L}_{\text{GRPO}} &= -\frac{1}{G}\sum_{i=1}^{G} \frac{1}{|y^{(i)}|} \sum_{t=1}^{|y^{(i)}|} \Big[ J_t^{(i)} - \beta \mathbb{D}_{\text{KL}}\bigl( \pi_\theta \parallel \pi_{\text{ref}} \bigr) \Big].
\end{aligned}
\end{equation}
Where $\rho_t^{(i)}$ denotes the per-token importance weight. While GRPO effectively drives exploration, it strictly evaluates outcome success and lacks the granular supervision required to optimize complex intermediate reasoning steps.

\subsection{Online Policy Self-Distillation (OPSD)}
To supply denser token-level supervision and follow the context-conditioned self-distillation formulation of OPSD~\citep{zhao2026self}, we obtain two next-token distributions from the same current policy parameters. At optimization step $k$, the student branch conditions on $s_t=(x,y_{<t})$, whereas the teacher branch additionally conditions on  $s_t^{+}=(x,p,y_{<t})$, where $p$ denotes the privileged information (PI) available exclusively during training. 
By restricting gradient flow solely to the student path during backpropagation, OPSD minimizes the per-token reverse KL divergence:
\begin{equation}
  \mathcal{L}_{\text{OPSD}}^{(t)} = \mathbb{D}_{\text{KL}} \bigl( \pi_{\theta_k}(\cdot \mid s_t) \parallel \pi_{\theta_k}(\cdot \mid s_t^{+}) \bigr).
\end{equation}
This homologous formulation inherently circumvents the vocabulary mismatch issues encountered in proprietary model distillation. However, standard OPSD typically derives the PI $p$ from the student's own correct rollouts, which fundamentally bounds the supervision quality by the student's existing reasoning limits.

%\vspace{-1mm}
\section{Methodology}\label{sec:methodology}
In this section, we first introduce the Structured JSON Protocol (Section~\ref{sec:json_protocol}) that normalizes privileged information and decouples cognitive strategies from proprietary model-specific linguistic patterns. 
We then detail the Multi-Agent System Generation Pipeline (Section~\ref{sec:mas_pipeline}), which automates the distillation of raw teacher trajectories into our compact protocol through a three-stage collaborative process with rigorous quality control. 
Finally, we present the Joint Training Objective (Section~\ref{sec:joint_training}) that combines a token-level OPSD distillation signal with a sparse GRPO reinforcement learning signal to effectively align the student model.

\begin{figure*}[t]
    \centering
    \includegraphics[width=0.95\textwidth]{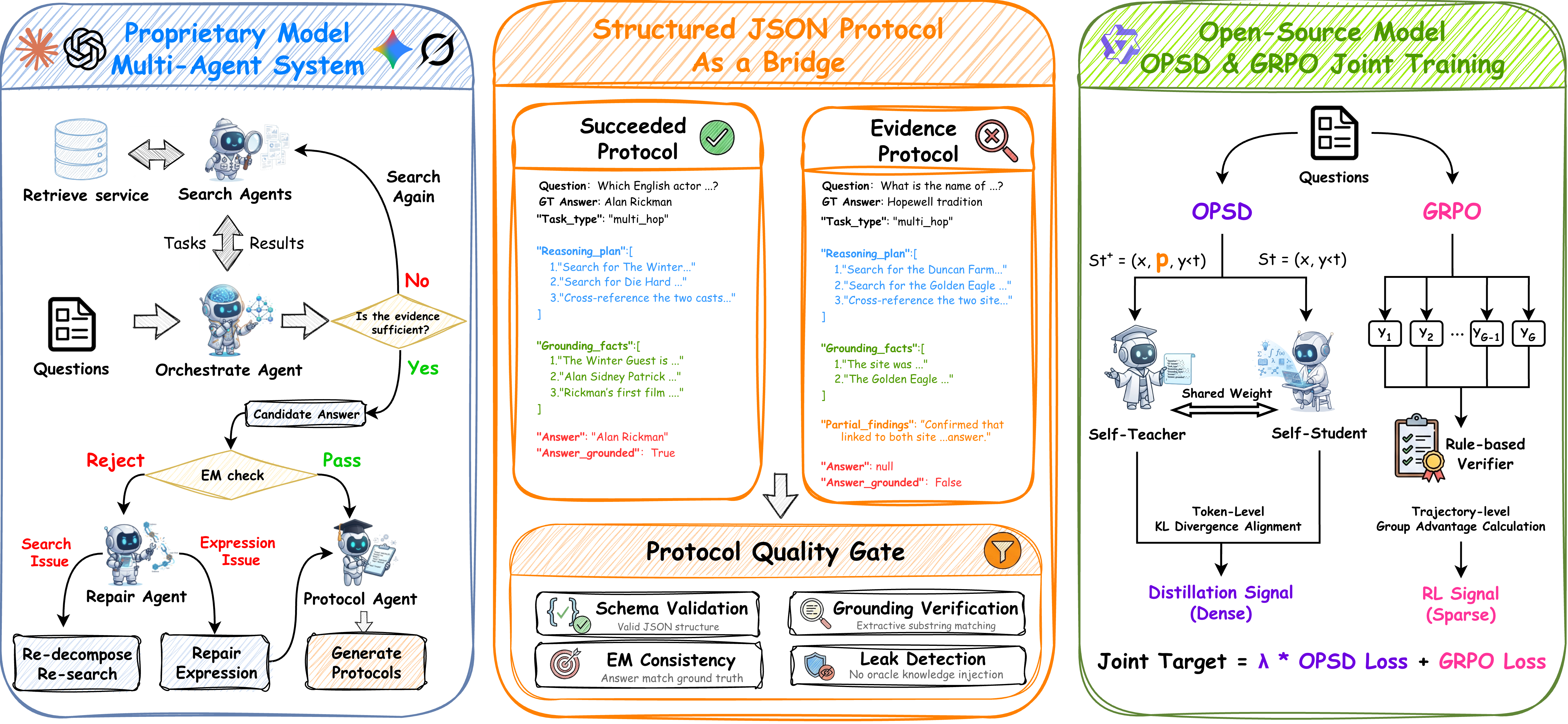} 
    \caption{Overview of the MAPD framework. (1) A proprietary-model-powered multi-agent system generates high-quality structured protocols. (2) The structured JSON protocol serves as a bridge between proprietary teacher and open-source student. (3) The student model internalizes the teacher's capability through joint training with protocols as privileged information.}
    \label{fig:framework}
\end{figure*}

\subsection{Structured JSON Protocol}\label{sec:json_protocol}
Directly utilizing free-form natural language trajectories as privileged information for OPSD can introduce a substantial representation mismatch or induce severe style drift and hallucinations, because they contain model-specific verbosity, formatting conventions, and linguistic patterns. 
To address this, we introduce a style-normalized Structured JSON Protocol as the fundamental unit of distillation, and this design reduces the student's direct exposure to teacher-specific language and provides traceable evidence for the privileged teacher branch.
Specifically, rather than extracting raw text, this protocol establishes a purified, traceable format (e.g., Example~\ref{protocolexmp:protocol_evidence}). Formally, the protocol $z$ encapsulates five essential dimensions of task:
\begin{itemize}
    \item \textbf{Task Type:} Determined by the proprietary expert, this field categorizes the query into \texttt{single\_hop}, \texttt{multi\_hop}, \texttt{comparison}, \texttt{others}, thereby dictating the overarching cognitive strategy.
    \item \textbf{Reasoning Plan:} An ordered array of actionable sub-goals that strictly decomposes the complex agentic task into tractable search and inference steps.
    \item \textbf{Grounding Facts:} A curated set of evidence extracted from the environment, which enforces faithful reasoning and actively mitigates parametric hallucinations.
    \item \textbf{Partial Findings:} An optional field recording intermediate discoveries when multiple search rounds fail to yield the correct answer.
    \item \textbf{Answer Verification:} The final extracted answer paired with a boolean flag (\texttt{answer\_grounded}), ensuring the conclusion is rigorously supported by the retrieved facts.
\end{itemize}

By formatting the protocol via a descriptive template $f(\cdot)$ such that the PI is defined as $p = f(z)$, we effectively decouple the core cognitive strategies from the proprietary model's surface linguistic patterns. This mechanism not only facilitates the transfer of high-quality reasoning capabilities into OPSD, but also effectively preserves the student's native token distribution.

\begin{figure*}[t]
\small
\begin{protocolexmp}{Representative Example of Structured Protocol --- Evidence Protocol}{protocol_evidence}
{}\textbf{Question:} \textit{What is the name of the culture cultivated in the area of Duncan Farm and Golden Eagle?} \par
\textbf{Ground-Truth Answer:} \textit{Hopewell tradition} \hfill \textcolor{answercolor}{\small(\emph{Search did not converge to this answer})} \par
\vspace{4pt}
\hrule
\vspace{6pt}
\texttt{\textbf{"task\_type":}} \texttt{\textbf{"multi\_hop"}} \par
\vspace{4pt}
\texttt{\textbf{"reasoning\_plan":}} \textcolor{plancolor}{[} \par
~~~~~\textcolor{plancolor}{1. ``Search for the Duncan Farm archaeological site to determine its geographic location and cultural period.''} \par
~~~~~\textcolor{plancolor}{2. ``Search for the Golden Eagle archaeological site to determine the culture it belongs to.''} \par
~~~~~\textcolor{plancolor}{3. ``Cross-reference two sites to find the name of the jointly present in the Duncan Farm and Golden Eagle area.''} 
\textcolor{plancolor}{]} \par
\vspace{4pt}
\texttt{\textbf{"grounding\_facts":}} \textcolor{factcolor}{[} \par
~~~~~\textcolor{factcolor}{1. ``The site was part of the Hopewell exchange system in Illinois and is the closest neighboring village site to the Golden Eagle regional transaction center, a major trade and social hub in the system.''} \par
~~~~~\textcolor{factcolor}{2. ``The Golden Eagle-Toppmeyer Site is a pre-Columbian archaeological site located near the confluence of the Illinois and Mississippi Rivers in Calhoun County, Illinois. The site is associated with the Havana Hopewell culture.''} 
\textcolor{factcolor}{]} \par
\vspace{4pt}
\texttt{\textbf{"partial\_findings":}} \textcolor{findingcolor}{``Confirmed that Duncan Farm belongs to the Woodland period and is part of the Hopewell exchange system, and that the Golden Eagle-Toppmeyer Site is explicitly associated with the Havana Hopewell culture; the culture jointly linked to both sites is the Havana Hopewell culture, but the search did not resolve whether the precise sense of `cultivated' points to a different answer.''} \par
\vspace{4pt}
\texttt{\textbf{"answer":}} \textcolor{answercolor}{\texttt{\textbf{null}}} \par
\texttt{\textbf{"answer\_grounded":}} \textcolor{answercolor}{\texttt{\textbf{false}}}
\end{protocolexmp}
\label{fig:example_protocol_evidence}
\end{figure*}

\subsection{Multi-Agent System Generation Pipeline}\label{sec:mas_pipeline}
While the structured JSON protocol mitigates style drift, reliably extracting such knowledge from a proprietary teacher remains challenging. We automate this via a three-stage MAS pipeline that distills the teacher's raw problem-solving into our compact protocol. Crucially, to prevent data leakage, ground-truth answers are strictly confined to offline synthesis and the privileged teacher context, remaining completely invisible to the student.

\paragraph{Stage~A: Multi-Agent Collaborative Search.}
Given a question $x$, the \textbf{Orchestrator agent} decomposes $x$ into sub-tasks with explicit dependency annotations. 
Subsequently, these sub-tasks are dispatched in parallel to independent \textbf{Searcher agents}. 
In response, each Searcher issues up to $K$ queries to a local Wikipedia corpus and compresses the retrieved passages into a concise finding. 
Once all findings within a topological level are collected, the Orchestrator synthesizes them; if the accumulated evidence remains insufficient, it formulates an additional round of sub-tasks (up to $M$ rounds). 
After the initial exploration pass, an exact match (EM) check evaluates whether the candidate answer matches the ground truth answer. 
If the check fails, a \textbf{Repair agent} leverages the ground-truth answer as diagnostic guidance to trace back the reasoning chain (The GT answer is never propagated into the exploration log and retrieval queries). 
It diagnoses the failure as either an \emph{expression} issue (correct evidence retrieved but answer poorly formatted) or a \emph{search issue}. It then triggers targeted re-decomposition and additional search operations accordingly, for up to $R$ repair iterations. Ultimately, the output of Stage~A comprises an original question, a candidate answer, a binary success label, and a comprehensive exploration log documenting every sub-task, query, retrieved passage, and finding.

\paragraph{Stage~B: Protocol Generation.}
To transform the exhaustive yet unstructured exploration log from Stage~A into our structured JSON protocol target, a \textbf{Protocolizer agent} is deployed. 
Conditioned on the success label, the protocolizer compiles this raw information into one of two structured JSON variants. For succeeded trajectories, it generates a \emph{Succeeded Protocol} that includes the fully verified answer. Conversely, if the search failed to converge, it generates an \emph{Evidence Protocol}; here, a definitive answer is omitted and replaced with a neutral summary of partial findings (e.g., ``Confirmed $X...$, but reliable evidence for $Y...$ is missing'').
To guarantee generation quality, two integrity constraints are strictly enforced: (i)~\emph{No Hindsight}: the reasoning\_plan must be formulated step-by-step, strictly prohibiting any mention of the final outcome or subsequent search results to prevent data leakage; and (ii)~\emph{Extractive Grounding}: the grounding\_facts must be strictly extractive, appearing as verbatim substrings within the retrieved passages.

\paragraph{Stage~C: Quality Gate.}
Before a generated protocol is admitted into the final training phase, it must pass a rigorous quality gate comprising four automated checks: (i)~\emph{Schema Validation}, ensuring a well-formed and valid JSON structure; (ii)~\emph{EM Consistency} (for succeeded protocol), confirming the extracted answer strictly matches the ground truth; (iii)~\emph{Grounding Verification}, enforcing extractive substring matching for all provided facts; and (iv)~\emph{Leak Detection}, guaranteeing no oracle knowledge is injected into the reasoning steps. Protocols failing any of these criteria are discarded, and the corresponding training falls back to a self-rollout baseline. 
Finally, manual review confirms the robustness of this pipeline, demonstrating a $99.33\%$ accuracy in yielding high-quality, strictly compliant structured JSON protocols (total $3,000$ instances across three representative models).

\subsection{Joint Training Objective}\label{sec:joint_training}
The overall training objective integrates a sparse, outcome-driven reinforcement learning signal with a dense distillation signal to achieve high-quality alignment:
\begin{equation}
\mathcal{L}(\theta) = \lambda_{OPSD} \cdot \mathcal{L}_{\text{OPSD}}(\theta) + \mathcal{L}_{\text{GRPO}}(\theta),
\end{equation}
where $\mathcal{L}_{\text{OPSD}}$ provides a token-level distillation signal derived from the structured protocol, which facilitates efficient step-by-step credit assignment. Conversely, $\mathcal{L}_{\text{GRPO}}$ provides a sparse RL signal via clipped policy gradients, optimizing the model toward the final environment reward. The hyperparameter $\lambda_{OPSD}$ dictates the relative strength of this distillation guidance (extensively ablated in Section~\ref{sec:exp_results}).

\section{Experiments and Analysis}
\subsection{Experimental Setup}\label{sec:setup}

\paragraph{Data Splits and Benchmarks.}
Following Search-R1~\citep{jin2025search}, our training uses only the NQ and HotpotQA train splits. 
Evaluation spans seven knowledge-intensive QA benchmarks, encompassing single-hop (NQ, TriviaQA, and PopQA) and multi-hop (HotpotQA, 2WikiMultihopQA, MuSiQue, and Bamboogle) reasoning tasks~\citep{kwiatkowski2019natural, min2019discrete, mallen2023not, yang2018hotpotqa, ho2020constructing, trivedi2022musique, press2023measuring}. 
Five of the seven evaluation sets remain entirely absent from training, serving as out-of-distribution tests; for the two shared benchmarks, we ensure no question-level overlap between train and test splits. Performance is reported as success rate (\%) via exact match against ground-truth answers.

\begin{table*}[t]
    \centering
    \resizebox{\textwidth}{!}{%
    \begin{tabular}{ll | ccccccc | >{\columncolor{gray!15}}c}
    \toprule
    \multirow{2}{*}{\textbf{Model}} & \multirow{2}{*}{\textbf{Method}} & \multicolumn{3}{c}{\textbf{Single-hop QA}} & \multicolumn{4}{c|}{\textbf{Multi-hop QA}} &
    \multicolumn{1}{c}{} \\
    \cmidrule(lr){3-5} \cmidrule(lr){6-9}
    & & \textbf{NQ} & \textbf{TriviaQA} & \textbf{PopQA} & \textbf{HotpotQA} & \textbf{2Wiki} & \textbf{MuSiQue} & \textbf{Bamboogle} & \textbf{Avg.} \\
    \midrule
    \multirow{6}{*}{\raisebox{-0.15em}{\includegraphics[height=1.2em]{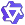}}~Qwen3-1.7B}
    & Vanilla & 29.9{\scriptsize$\pm$0.3} & 46.4{\scriptsize$\pm$0.2} & 39.1{\scriptsize$\pm$0.1} & 23.9{\scriptsize$\pm$0.2} & 18.8{\scriptsize$\pm$0.1} & 4.9{\scriptsize$\pm$0.3} &
  16.8{\scriptsize$\pm$1.1} & 25.7{\scriptsize$\pm$0.1} \\
    & OPSD & 4.2{\scriptsize$\pm$0.1} & 8.3{\scriptsize$\pm$0.1} & 4.6{\scriptsize$\pm$0.1} & 6.6{\scriptsize$\pm$0.1} & 15.3{\scriptsize$\pm$0.1} & 0.7{\scriptsize$\pm$0.1} &
  1.6{\scriptsize$\pm$0.5} & 5.9{\scriptsize$\pm$0.1} \\
    & GRPO & 36.9{\scriptsize$\pm$0.4} & 54.6{\scriptsize$\pm$0.3} & 43.1{\scriptsize$\pm$0.1} & 29.8{\scriptsize$\pm$0.3} & 27.5{\scriptsize$\pm$0.2} & 6.8{\scriptsize$\pm$0.4} &
  22.4{\scriptsize$\pm$1.6} & 31.6{\scriptsize$\pm$0.2} \\
    & GRPO+OPSD & 37.0{\scriptsize$\pm$0.3} & 54.6{\scriptsize$\pm$0.2} & 41.4{\scriptsize$\pm$0.1} & 29.9{\scriptsize$\pm$0.3} & 23.3{\scriptsize$\pm$0.2} & 6.5{\scriptsize$\pm$0.3} &
  20.8{\scriptsize$\pm$1.6} & 30.5{\scriptsize$\pm$0.2} \\
    & SDAR & \underline{43.4}{\scriptsize$\pm$0.4} & \underline{58.1}{\scriptsize$\pm$0.3} & \underline{47.6}{\scriptsize$\pm$0.2} & \underline{37.0}{\scriptsize$\pm$0.3} &
  \underline{34.3}{\scriptsize$\pm$0.3} & \underline{10.6}{\scriptsize$\pm$0.5} & \underline{32.0}{\scriptsize$\pm$1.6} & \underline{37.6}{\scriptsize$\pm$0.3} \\
    & \textbf{MAPD(Ours)} & \textbf{45.1}{\scriptsize$\pm$0.2} & \textbf{58.6}{\scriptsize$\pm$0.2} & \textbf{48.6}{\scriptsize$\pm$0.1} & \textbf{39.0}{\scriptsize$\pm$0.2} &
  \textbf{36.2}{\scriptsize$\pm$0.2} & \textbf{11.2}{\scriptsize$\pm$0.3} & \textbf{36.8}{\scriptsize$\pm$1.1} & \textbf{39.4}{\scriptsize$\pm$0.1} \\
    \midrule
    \multicolumn{2}{c|}
    {\textcolor{green!60!black}{\textbf{Improvement (\%)}}} & \textcolor{green!60!black}{3.9\%\,$\uparrow$} & \textcolor{green!60!black}{0.9\%\,$\uparrow$} & \textcolor{green!60!black}{2.1\%\,$\uparrow$} & \textcolor{green!60!black}{5.4\%\,$\uparrow$} & \textcolor{green!60!black}{5.5\%\,$\uparrow$} & \textcolor{green!60!black}{5.7\%\,$\uparrow$} &
    \textcolor{green!60!black}{15.0\%\,$\uparrow$} & \textcolor{green!60!black}{4.8\%\,$\uparrow$} \\
    \midrule
    \multirow{6}{*}{\raisebox{-0.15em}{\includegraphics[height=1.2em]{figures/qwen.pdf}}~Qwen3-4B}
    & Vanilla & 29.3{\scriptsize$\pm$0.3} & 48.0{\scriptsize$\pm$0.2} & 34.9{\scriptsize$\pm$0.1} & 26.5{\scriptsize$\pm$0.2} & 30.9{\scriptsize$\pm$0.2} & 6.1{\scriptsize$\pm$0.3} &
  25.6{\scriptsize$\pm$1.6} & 28.8{\scriptsize$\pm$0.2} \\
    & OPSD & 16.2{\scriptsize$\pm$0.2} & 36.7{\scriptsize$\pm$0.1} & 23.1{\scriptsize$\pm$0.1} & 20.5{\scriptsize$\pm$0.2} & 24.6{\scriptsize$\pm$0.1} & 5.7{\scriptsize$\pm$0.2} &
  19.2{\scriptsize$\pm$1.1} & 20.9{\scriptsize$\pm$0.1} \\
    & GRPO & 40.4{\scriptsize$\pm$0.3} & 61.0{\scriptsize$\pm$0.2} & 43.5{\scriptsize$\pm$0.1} & 36.7{\scriptsize$\pm$0.3} & 37.3{\scriptsize$\pm$0.2} & 10.3{\scriptsize$\pm$0.4} &
  32.8{\scriptsize$\pm$1.6} & 37.4{\scriptsize$\pm$0.2} \\
    & GRPO+OPSD & 36.3{\scriptsize$\pm$0.3} & 57.9{\scriptsize$\pm$0.2} & 43.1{\scriptsize$\pm$0.1} & 37.7{\scriptsize$\pm$0.3} & 36.7{\scriptsize$\pm$0.2} & 12.7{\scriptsize$\pm$0.3} &
  44.0{\scriptsize$\pm$1.6} & 38.3{\scriptsize$\pm$0.2} \\
    & SDAR & \underline{46.1}{\scriptsize$\pm$0.4} & \underline{62.4}{\scriptsize$\pm$0.2} & \underline{48.5}{\scriptsize$\pm$0.2} & \underline{43.3}{\scriptsize$\pm$0.3} &
  \underline{43.1}{\scriptsize$\pm$0.3} & \underline{13.1}{\scriptsize$\pm$0.5} & \underline{44.8}{\scriptsize$\pm$1.6} & \underline{43.0}{\scriptsize$\pm$0.3} \\
    & \textbf{MAPD(Ours)} & \textbf{47.7}{\scriptsize$\pm$0.2} & \textbf{62.5}{\scriptsize$\pm$0.2} & \textbf{49.4}{\scriptsize$\pm$0.1} & \textbf{44.3}{\scriptsize$\pm$0.2} &
  \textbf{45.7}{\scriptsize$\pm$0.2} & \textbf{14.0}{\scriptsize$\pm$0.3} & \textbf{47.2}{\scriptsize$\pm$1.1} & \textbf{44.4}{\scriptsize$\pm$0.1} \\
    \midrule
    \multicolumn{2}{c|}
    {\textcolor{green!60!black}{\textbf{Improvement (\%)}}} & \textcolor{green!60!black}{3.5\%\,$\uparrow$} & \textcolor{green!60!black}{0.2\%\,$\uparrow$} & \textcolor{green!60!black}{1.9\%\,$\uparrow$} & \textcolor{green!60!black}{2.3\%\,$\uparrow$} & \textcolor{green!60!black}{6.0\%\,$\uparrow$} & \textcolor{green!60!black}{6.9\%\,$\uparrow$} &
    \textcolor{green!60!black}{5.4\%\,$\uparrow$} & \textcolor{green!60!black}{3.3\%\,$\uparrow$} \\
    \bottomrule
    \end{tabular}%
    }
    \caption{Results are reported as success rate (\%) (mean $\pm$ std across five independent evaluation seeds from a single training run). \textbf{Avg.} denotes the macro-averaged success rate. Improvement (\%) is the relative gain of MAPD over the SDAR baseline. \textbf{Bold}: best performance; \underline{underline}: second best.}
    \label{tab:agentic_search_results}
\end{table*}

\paragraph{Baselines.}
To rigorously assess the effectiveness of our approach, we compare against a spectrum of baselines, ranging from pure alignment strategies to advanced hybrid methods:
(1)~\textbf{Vanilla}: The pretrained base model evaluated zero-shot without any further alignment;
(2)~\textbf{OPSD}~\citep{zhao2026self}: Pure online policy self-distillation, which utilizes the student's own successful rollouts as privileged information;
(3)~\textbf{GRPO}~\citep{shao2024deepseekmath}: Pure outcome-based reinforcement learning, relying solely on sparse environment rewards without any dense distillation guidance;
(4)~\textbf{GRPO+OPSD}: A naive hybrid approach combining RL with online policy self-distillation, using successful rollouts as privileged information;
(5)~\textbf{SDAR}~\citep{lu2026self}: An advanced hybrid baseline combining RL with token-level gated distillation, utilizing skill-generated content as privileged information;
(6)~\textbf{MAPD (Ours)}: Our proposed Multi-Agent Protocol Distillation framework, integrating RL with dense, token-level distillation guided by the structured MAS protocol without gating mechanisms.

\paragraph{Implementation Details.}
For the student policy, we train two open-source models, Qwen3-1.7B and Qwen3-4B~\citep{yang2025qwen3}, both initialized from their respective pretrained base checkpoints. During joint training, we utilize a group size of $n\!=\!8$ rollouts per prompt, a batch size of $128$, a maximum prompt length of $4096$, and a maximum response length of $512$ tokens. The learning rate is set to $1\!\times\!10^{-6}$ with a $10\%$ linear warmup. Training runs for $200$ steps across 8 GPUs. The agentic environment allows up to $4$ interaction turns, utilizing a local Wikipedia corpus (wiki-18) as the retrieval backend, returning the top-3 passages per query.

\paragraph{MAS Pipeline Configuration.}
For protocol synthesis, the MAS agents (Orchestrator, Searcher, Repair, and Protocolizer) utilize Claude-Opus-4.6 as the default backbone LLM. We also explore GPT-5.5 and Gemini-3.1-Pro as alternative teacher sources (ablated in Section~\ref{sec:exp_results}). The MAS parameters are constrained as follows: the Orchestrator decomposes each query into a maximum of $4$ sub-questions over at most $2$ rounds; each Searcher issues up to $3$ retrieval queries; and the Repair agent is permitted up to $2$ diagnostic rounds. To ensure training efficiency, all protocols are pre-synthesized offline and cached, entirely decoupling the teacher's generation latency from the student's RL loop.

\subsection{Results and Analysis}~\label{sec:exp_results}

\paragraph{Main Result.}
As shown in Table~\ref{tab:agentic_search_results}, our proposed MAPD framework consistently outperforms all baseline methods across the seven benchmarks, achieving average success rates of $39.4\%$ on Qwen3-1.7B and $44.4\%$ on Qwen3-4B. Compared to the strongest baseline SDAR, MAPD yields relative average improvements of $4.8\%$ and $3.3\%$, respectively. 
Crucially, these performance gains are more pronounced on complex multi-hop QA tasks. For Qwen3-1.7B, the average relative gain on multi-hop benchmarks reaches 7.9\%, substantially exceeding the 2.3\% on single-hop benchmarks. A similar pattern holds for Qwen3-4B (5.1\% vs.\ 1.8\%). 
These larger relative gains on multi-hop benchmarks are consistent with the intended role of multi-agent decomposition and evidence aggregation, as the MAS pipeline in MAPD systematically breaks down complex queries, synthesizes evidence across multiple retrieval steps, and distills these strategies into the student policy via our structured protocol, thereby raising its upper bound for complex reasoning. 
Furthermore, we observe a catastrophic performance collapse when applying pure OPSD to the models under our agentic training configuration (dropping to 5.9\% on Qwen3-1.7B and 20.9\% on Qwen3-4B). 
This indicates that dense self-distillation is insufficient for agentic search and motivates the use of external PI.
Lacking external information gain, self-rollouts merely drive the model toward pathologically long outputs——as evidenced by the length clip ratio surging from $5\%$ to approximately $74\%$ during training, a collapse that underscores the necessity of our MAS protocol.

\begin{table}[t]
\centering
\small
\setlength{\tabcolsep}{7pt}
\begin{tabular}{@{}l ccc cc@{}}
\toprule
\multirow{2}{*}{\textbf{Method}} & \multirow{2}{*}{\textbf{PM}} & \multirow{2}{*}{\textbf{SP}} & \multirow{2}{*}{\textbf{MAS}} & \multicolumn{2}{c}{\textbf{Avg.}} \\
\cmidrule(l){5-6}
& & & & \textbf{Qwen3-1.7B} & \textbf{Qwen3-4B} \\
\midrule
GRPO+OPSD       & \ding{55} & \ding{55} & \ding{55} & 30.5 & 38.3 \\
SDAR             & \ding{55} & \ding{55} & \ding{55} & 37.6 & 43.0 \\
\midrule
Raw trajectory (single PM)    & \ding{51} & \ding{55} & \ding{55} & 30.1 & 37.3 \\
Raw trajectory (MAS)           & \ding{51} & \ding{55} & \ding{51} & 29.2 & 36.1 \\
Protocol (single PM)          & \ding{51} & \ding{51} & \ding{55} & 37.1 & 42.9 \\
\midrule
\textbf{MAPD (Ours)} & \ding{51} & \ding{51} & \ding{51} & \textbf{39.4} & \textbf{44.4} \\
\bottomrule
\end{tabular}
\caption{Component ablation study of MAPD. PM = Proprietary Model; SP = Structured Protocol; MAS = Multi-Agent System.}
\label{tab:component_ablation}
\end{table}

\paragraph{Ablation Study.}
We systematically ablate the Proprietary Model (PM), Structured Protocol (SP), and Multi-Agent System (MAS) to assess the contribution of each component, results are reported in Table~\ref{tab:component_ablation}.

\textbf{\textit{(1) Structured protocol is essential for cross-model distillation.}}
The most striking finding is that directly distilling from a proprietary model's raw natural-language trajectories (\emph{w/o SP\&MAS}) performs worse than baselines using no proprietary model at all (\emph{GRPO+OPSD}). 
By comparing \emph{w/o SP\&MAS} with \emph{w/o MAS}, we isolate the impact of the representation format under a single-teacher setting: replacing raw text with our structured protocol drives an observed gain of 7.0 and 5.6 points, respectively.
This is consistent with the style drift and distribution gap bottlenecks discussed in Section~1: without structured decoupling, the student imitates the teacher's surface patterns (e.g., verbose reasoning chains, idiosyncratic phrasing), which conflicts with its own learned distribution and actively degrades performance.

\textbf{\textit{(2) MAS cannot replace structured protocol.}}
Adding the MAS pipeline without the structured protocol (\emph{w/o SP}: 29.2\%/36.1\%) does not improve performance and is slightly worse, because the MAS produces richer but also noisier natural-language trajectories that exacerbate the style-transfer problem. 
Conversely, applying the structured protocol generated by single proprietary agent (\emph{w/o MAS}: 37.1\%/42.9\%) recovers a relative gain, showing that the JSON protocol alone already alleviates style contamination. However, this single-agent protocol still underperforms SDAR, indicating that a single proprietary model call cannot match the retrieval depth and error recovery of the full MAS pipeline.

\textbf{\textit{(3) MAS pipeline improves the quality of protocol generation.}}
The full MAPD configuration performs best, suggesting that the structured representation and the MAS pipeline provide complementary benefits in the evaluated setting. Specifically, SP ensures that the distilled signal largely mitigates style drift and remains learnable, while MAS ensures that the signal is factually accurate, reasoning-complete and traceable.
Compared with the single-agent protocol without MAS, this full protocol delivers a significantly higher-quality training signal that bridges the gap between proprietary and open-source models. 
Moreover, our results demonstrate that the primary bottleneck in distillation lies in PI quality rather than training mechanisms: with sufficiently clean, style-decoupled protocols, a simple token-level KL objective achieves performance comparable to gated approaches (SDAR).

\begin{table*}[t]
  \centering
  \resizebox{\textwidth}{!}{%
  \begin{tabular}{ll c | ccccccc | >{\columncolor{gray!15}}c}
  \toprule
  \multirow{2}{*}{\textbf{Proprietary Teacher}} & \multirow{2}{*}{\textbf{Model}} & \multirow{2}{*}{$\boldsymbol{\lambda}$} & \multicolumn{3}{c}{\textbf{Single-hop QA}} &
  \multicolumn{4}{c|}{\textbf{Multi-hop QA}} & \multicolumn{1}{c}{} \\
  \cmidrule(lr){4-6} \cmidrule(lr){7-10}
  & & & \textbf{NQ} & \textbf{TriviaQA} & \textbf{PopQA} & \textbf{HotpotQA} & \textbf{2Wiki} & \textbf{MuSiQue} & \textbf{Bamboogle} & \textbf{Avg.} \\
  \midrule
  \multirow{6}{*}{\raisebox{-0.15em}{\includegraphics[height=1.2em]{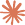}}~Claude-Opus-4.6}
  & \multirow{3}{*}{\raisebox{-0.15em}{\includegraphics[height=1.2em]{figures/qwen.pdf}}~Qwen3-1.7B}
  & 0.01 & 44.1 & 58.0 & 47.5 & 37.8 & 33.5 & 10.9 & 28.0 & 37.1 \\
  & & 0.05 & \textbf{45.1} & \textbf{58.6} & 48.6 & \textbf{39.0} & 36.2 & \textbf{11.2} & \textbf{36.8} & \textbf{39.4} \\
  & & 0.10 & 43.7 & 58.2 & \textbf{49.1} & 38.2 & \textbf{36.6} & 10.2 & 36.0 & 38.9 \\
  \cmidrule(l){2-11}
  & \multirow{3}{*}{\raisebox{-0.15em}{\includegraphics[height=1.2em]{figures/qwen.pdf}}~Qwen3-4B}
  & 0.01 & 46.3 & 63.0 & 48.3 & 42.5 & 37.0 & 13.9 & 43.2 & 42.0 \\
  & & 0.05 & \textbf{47.7} & 62.5 & 49.4 & \textbf{44.3} & \textbf{45.7} & \textbf{14.0} & \textbf{47.2} & \textbf{44.4} \\
  & & 0.10 & 47.4 & \textbf{63.7} & \textbf{50.1} & 42.3 & 38.4 & 11.1 & 42.4 & 42.2 \\
  \midrule
  \multirow{6}{*}{\raisebox{-0.15em}{\includegraphics[height=1.2em]{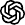}}~GPT-5.5}
  & \multirow{3}{*}{\raisebox{-0.15em}{\includegraphics[height=1.2em]{figures/qwen.pdf}}~Qwen3-1.7B}
  & 0.01 & \textbf{43.5} & 58.6 & \textbf{49.1} & 38.4 & 35.5 & \textbf{11.2} & \textbf{36.0} & 38.9 \\
  & & 0.05 & \textbf{43.5} & \textbf{59.7} & 48.6 & \textbf{38.7} & \textbf{38.0} & 9.3 & 35.2 & \textbf{39.0} \\
  & & 0.10 & 43.3 & 58.4 & 48.6 & 36.4 & 31.9 & 9.2 & 32.8 & 37.2 \\
  \cmidrule(l){2-11}
  & \multirow{3}{*}{\raisebox{-0.15em}{\includegraphics[height=1.2em]{figures/qwen.pdf}}~Qwen3-4B}
  & 0.01 & 47.4 & 62.2 & 48.7 & 43.5 & \textbf{46.0} & 14.3 & 46.4 & 44.1 \\
  & & 0.05 & \textbf{48.3} & 62.9 & \textbf{50.6} & \textbf{44.2} & 44.6 & \textbf{14.6} & \textbf{47.2} & \textbf{44.6} \\
  & & 0.10 & 47.4 & \textbf{63.7} & 50.3 & 43.4 & 43.1 & 12.6 & 44.0 & 43.5 \\
  \midrule
  \multirow{6}{*}{\raisebox{-0.15em}{\includegraphics[height=1.2em]{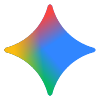}}~Gemini-3.1-Pro}
  & \multirow{3}{*}{\raisebox{-0.15em}{\includegraphics[height=1.2em]{figures/qwen.pdf}}~Qwen3-1.7B}
  & 0.01 & 42.8 & 58.2 & \textbf{48.9} & 37.4 & 32.4 & 9.9 & 33.6 & 37.6 \\
  & & 0.05 & \textbf{43.2} & \textbf{58.9} & 47.5 & \textbf{37.7} & \textbf{36.3} & 9.1 & 32.8 & \textbf{37.9} \\
  & & 0.10 & 42.5 & 58.3 & 47.9 & \textbf{37.7} & 32.6 & \textbf{10.5} & \textbf{35.2} & 37.8 \\
  \cmidrule(l){2-11}
  & \multirow{3}{*}{\raisebox{-0.15em}{\includegraphics[height=1.2em]{figures/qwen.pdf}}~Qwen3-4B}
  & 0.01 & 48.8 & \textbf{63.7} & 47.9 & \textbf{44.8} & \textbf{45.2} & 13.0 & 43.2 & 43.8 \\
  & & 0.05 & \textbf{48.9} & 62.8 & \textbf{50.8} & 43.8 & 39.1 & \textbf{14.0} & \textbf{48.8} & \textbf{44.0} \\
  & & 0.10 & 47.4 & 62.5 & 49.3 & 43.0 & 42.3 & 12.8 & 43.2 & 42.9 \\
  \bottomrule
  \end{tabular}%
  }
  \caption{Effect of three proprietary models as privileged information source and OPSD loss weight $\lambda$. Results are reported as Success Rate (\%). \textbf{Bold}: best performance.}
  \label{tab:pi_source_ablation}
\end{table*}

\begin{figure}[t]
    \centering
    \begin{minipage}{0.45\textwidth}
        \centering
        \includegraphics[width=\linewidth]{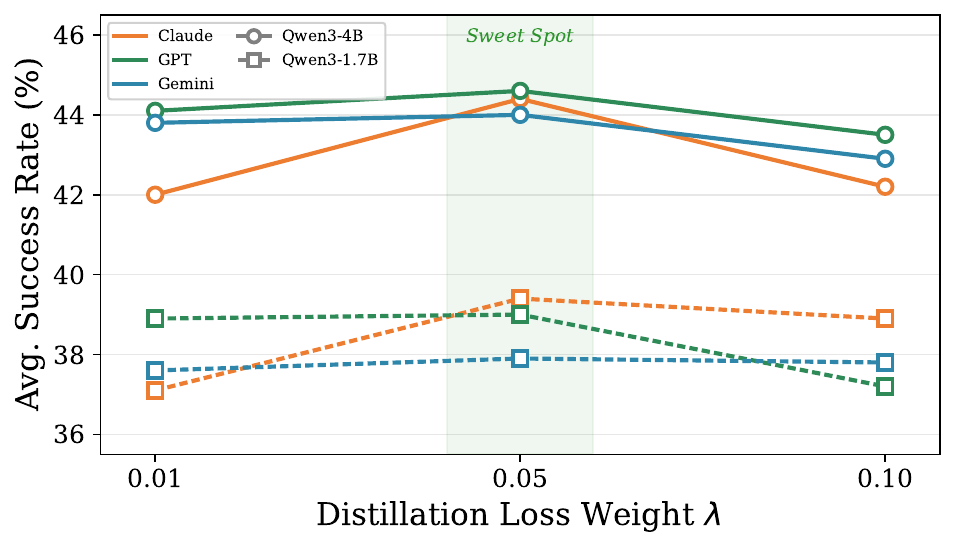}
        \caption{The relationship between the OPSD loss weight $\lambda$ and Avg success rate (\%). $\lambda\!=\!0.05$ balances distillation strength and policy stability, achieving the best Avg on both model scales.}
        \label{fig:lambda_sensitivity}
    \end{minipage}
    \hfill
    \begin{minipage}{0.52\textwidth}
        \centering
        \includegraphics[width=\linewidth]{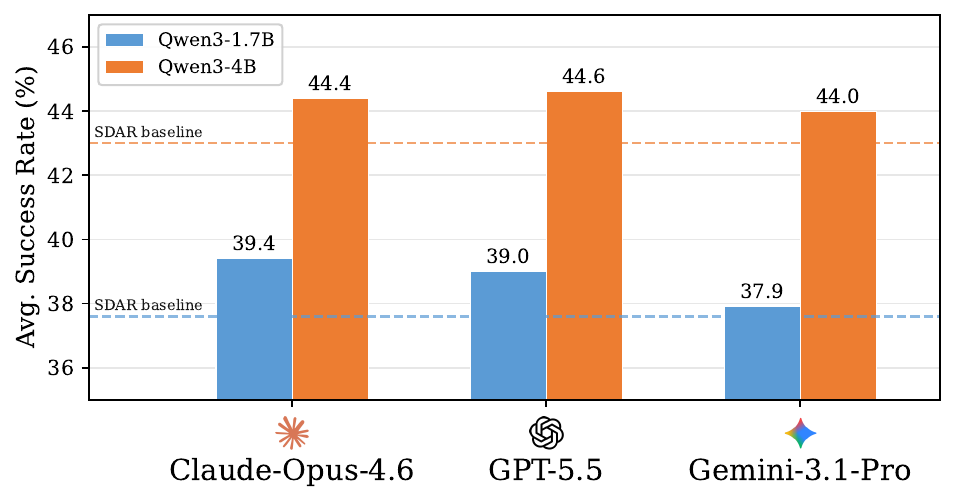}
        \caption{Performance with three top-tier proprietary teacher backbones. All configurations use the same MAPD architecture and training settings without retuning the MAS pipeline.}
        \label{fig:teacher_agnostic}
    \end{minipage}
\end{figure}

\paragraph{Effect of the OPSD Loss Weight.}
Table~\ref{tab:pi_source_ablation} and Figure~\ref{fig:lambda_sensitivity} summarizes the sensitivity of our framework to the OPSD loss weight $\lambda\!\in\!\{0.01,0.05,0.1\}$ in the joint objective. Across both model scales, $\lambda\!=\!0.05$ consistently achieves the best among the tested values (39.4\% on 1.7B and 44.4\% on 4B), revealing a sweet-spot associated with two observed patterns at the extremes. Below we analyze these in detail.

\textbf{\textit{(1) Weak distillation signal}} ($\lambda\!=\!0.01$). A small OPSD weight provides only a weak distillation signal relative to the dominant GRPO policy-gradient objective. Training logs on Claude-Opus-4.6 reveal that the teacher-student gap collapses prematurely by step~100, indicating that the student quickly exhausts the available supervision. However, the KL divergence from the reference policy remains low (0.24 on 1.7B), suggesting that the model does not simply revert to the reference behavior. Despite the early disappearance of the distillation signal, this configuration still achieves a clear improvement over GRPO alone (37.1\% vs.\ 31.6\% on 1.7B), implying that even a short-lived OPSD term can provide beneficial guidance during the initial phase of training.

\textbf{\textit{(2) Excessive distillation pressure}} ($\lambda\!=\!0.1$). A large OPSD weight pushes the student aggressively toward the teacher's distribution, which inflates the KL divergence from the reference policy (1.06 on 4B, nearly $4\times$ that of $\lambda\!=\!0.01$). More importantly, we observe a behavioral degeneration specific to agentic search: on 4B the mean response length collapses from 135 tokens to roughly 42 tokens while the number of tool calls saturates at the maximum of 3.0 per episode. This pattern suggests that excessive distillation pressure drives the model toward a ``retrieve-without-reasoning'' shortcut--issuing retrieval calls with minimal intermediate reasoning. 
As a result, single-hop accuracy slightly improves (e.g., PopQA rises from 49.4\% to 50.1\%, TriviaQA from 62.5\% to 63.7\% on 4B), but performance on multi-hop tasks drops sharply because coherent chain-of-thought across hops is essential (2WikiMultihopQA falls from 45.7\% to 38.4\%, MuSiQue from 14.0\% to 11.1\%). This trade-off reveals that over-regularization erodes the model's capacity for multi-step reasoning, ultimately harming the very capability that agentic search relies on.

Taken together, these results highlight a fundamental tension in agentic distillation: too little distillation provides insufficient guidance, while too much forces the student into shallow, retrieval-heavy behaviors. $\lambda\!=\!0.05$ strikes a practical balance, maintaining a sustained distillation signal throughout training without destabilizing the policy, thereby fostering a robust balance between efficient retrieval and rigorous reasoning.

% \begin{figure}[t]
%     \centering
%     \includegraphics[width=0.8\textwidth]{figures/fig_lambda_sensitivity.pdf}
%     \caption{Sensitivity to OPSD loss weight $\lambda$. $\lambda\!=\!0.05$ balances distillation strength and policy stability, achieving the best Avg on both model scales.}
%     \label{fig:lambda_sensitivity}
% \end{figure}

\paragraph{Transfer Across Proprietary Teacher Backbones.}
A key advantage of MAPD is transferability across different proprietary models. Because the structured JSON protocol successfully decouples the teacher's core reasoning strategies from its surface stylistic artifacts, any sufficiently capable model can serve as the teacher backbone. To validate this flexibility, we evaluate whether the framework can seamlessly utilize protocols generated by three top-tier commercial models without requiring any architectural modifications or teacher-specific hyperparameter tuning---Claude-Opus-4.6, GPT-5.5, and Gemini-3.1-Pro. 
Results are shown in Table~\ref{tab:pi_source_ablation} and Figure~\ref{fig:teacher_agnostic}. 
All three configurations produce positive observed gains over SDAR on both student scales: Claude-Opus-4.6 (39.4\%/44.4\%), GPT-5.5 (39.0\%/44.6\%), and Gemini-3.1-Pro (37.9\%/44.0\%). This narrow variance across teachers (within 2 points) confirms that our protocol successfully abstracts away teacher-specific idiosyncrasies, isolating semantic reasoning from the source model. Consequently, practitioners can seamlessly swap APIs to optimize cost-latency tradeoffs without retuning the pipeline.

\paragraph{Cost Analysis.}
To quantify the overhead of protocol synthesis, we report the generation cost using Gemini-3.1-Pro as an illustrative example. The MAS pipeline processes 25,600 training instances, producing 25,584 valid protocols that pass the quality gate (99.94\% yield). On average, the pipeline invokes the teacher LLM ${\sim}6.3$ times per instance, consuming approximately 12.5K tokens per instance. The amortized cost is \$0.057 per instance, yielding a total one-time generation cost of approximately \$1,454 for the entire training set. Since all protocols are pre-synthesized and cached, this cost is incurred only once.

%\vspace{-1mm}
\section{Related Work}\label{sec:related_work}
\paragraph{LLM Reasoning and Agentic Search.}
The paradigm of LLM reasoning has recently shifted from static, single-turn tasks toward dynamic, multi-turn agentic tasks that require sustained interaction with external environments~\citep{liu2026amo, liu2026general365, dong2025agentic2, mai2025agent, lv2026physagent, jiang2026beyond, jiang2025kore, jiang2026mined}. 
As a critical extension of this reasoning paradigm, agentic search interleaves multi-step logical deduction with environment-augmented retrieval to tackle knowledge-intensive tasks. 
Recent frameworks have formalized this as an RL optimization problem. For instance, WebRL~\citep{qi2025webrl} employs self-evolving RL with process-level rewards, Search-R1~\citep{jin2025search} leverages classic RL algorithms for multi-turn retrieval, and RAGEN~\citep{wang2025ragen} optimizes retrieval-augmented reasoning chains via trajectory-based rewards. However, these policies remain predominantly driven by sparse, outcome-only signals (e.g., terminal answer accuracy), which fundamentally struggle to provide granular credit assignment for intermediate planning, searching, and diagnostic steps. 
To resolve this optimization bottleneck, our MAPD framework introduces a joint training paradigm that integrates outcome-driven RL signal with a dense distillation signal.

\paragraph{Knowledge Distillation from Proprietary Models.}
Transferring capabilities from proprietary LLMs to smaller open-source models has become a dominant paradigm. Classical knowledge distillation~\citep{hinton2015distilling} relies on matching output probability distributions, a mechanism fundamentally precluded by the opaque nature of black-box APIs and the mismatch between heterogeneous tokenizers. Consequently, contemporary approaches resort to text-level imitation, distilling the natural language reasoning traces or Chain-of-Thought (CoT) generated by proprietary teachers~\citep{hsieh2023distilling, mukherjee2023orca, xu2025magpie}. However, as highlighted by recent studies~\citep{gudibande2023false}, directly mimicking unconstrained teacher trajectories forces the student to overfit to the teacher's verbose linguistic patterns and authoritative tone. This capacity mismatch induces severe style drift and exacerbates hallucinations. 
Our MAPD framework extracts the teacher's semantic strategies into a style-normalized structured JSON protocol to decouple the core reasoning logic from the proprietary linguistic shell, enabling safe online self-distillation.

\paragraph{Multi-Agent System for Complex Reasoning.}
LLM-based MAS have achieved remarkable success in complex reasoning and open-ended problem-solving~\citep{wu2023autogen, hong2024metagpt, lei2024macm}. In retrieval-augmented scenarios, multi-agent collaboration can significantly enhance recall on multi-step questions through parallel sub-query orchestration and iterative debate~\citep{du2023improving, talebirad2023multi, wang2026mad}. However, deploying these collaborative swarms at inference time incurs prohibitive computational overhead, severe latency, and massive API costs, rendering them highly impractical for real-time applications. Instead of relying on MAS during test-time interactions, MAPD utilizes the agent swarm exclusively as an offline training-time knowledge synthesizer, distilling the swarm's collaborative problem-solving strategies into a single student model via our structured protocol.

%\vspace{-1mm}
\section{Conclusion}\label{sec:conclusion}
In this paper, we introduce Multi-Agent Protocol Distillation (MAPD), a novel joint distillation and RL framework designed to bridge the heterogeneous distribution gap between proprietary teachers and open-source students in agentic search. 
To overcome the severe style drift and verbosity collapse inherent in raw trajectory imitation, MAPD elegantly decouples core cognitive strategies from idiosyncratic linguistic shells via a style-normalized, structured JSON protocol. Synthesized exclusively offline by a robust multi-agent system, this protocol serves as privileged information to provide dense guidance that seamlessly complements sparse outcome-driven RL without incurring any inference overhead. 
Extensive evaluations across seven knowledge-intensive QA benchmarks demonstrate that MAPD not only consistently outperforms all evaluated baselines, but also generalizes seamlessly across different proprietary teachers without retuning. 
By isolating semantic reasoning from idiosyncratic stylistic artifacts, MAPD unlocks a safer and more efficient pathway for heterogenes models alignment.
We expect that the structural decoupling pioneered by MAPD will serve as a catalyst for future research, inspiring the community to push beyond the reliance on proprietary models and toward the development of fully autonomous, self-improving agentic frameworks.

\newpage

\bibliography{conference}

@article{kwiatkowski2019natural,
  title={Natural questions: a benchmark for question answering research},
  author={Kwiatkowski, Tom and Palomaki, Jennimaria and Redfield, Olivia and Collins, Michael and Parikh, Ankur and Alberti, Chris and Epstein, Danielle and Polosukhin, Illia and Devlin, Jacob and Lee, Kenton and others},
  journal={Transactions of the Association for Computational Linguistics},
  volume={7},
  pages={453--466},
  year={2019},
  publisher={MIT Press One Rogers Street, Cambridge, MA 02142-1209, USA journals-info~…}
}

@inproceedings{min2019discrete,
  title={A discrete hard EM approach for weakly supervised question answering},
  author={Min, Sewon and Chen, Danqi and Hajishirzi, Hannaneh and Zettlemoyer, Luke},
  booktitle={Proceedings of the 2019 conference on empirical methods in natural language processing and the 9th international joint conference on natural language processing (EMNLP-IJCNLP)},
  pages={2851--2864},
  year={2019}
}

@article{luo2025globalrag,
  title={GlobalRAG: Enhancing Global Reasoning in Multi-hop Question Answering via Reinforcement Learning},
  author={Luo, Jinchang and Cheng, Mingquan and Wan, Fan and Li, Ni and Xia, Xiaoling and Tian, Shuangshuang and Bian, Tingcheng and Wang, Haiwei and Fu, Haohuan and Tao, Yan},
  journal={arXiv preprint arXiv:2510.20548},
  year={2025}
}

@article{liang2026search,
  title={Search-E1: Self-Distillation Drives Self-Evolution in Search-Augmented Reasoning},
  author={Liang, Zihan and Ma, Yufei and Chen, Ben and Qian, Zhipeng and Zhang, Xuxin and Dai, Huangyu and Mao, Lingtao},
  journal={arXiv preprint arXiv:2605.22511},
  year={2026}
}

@inproceedings{mallen2023not,
  title={When not to trust language models: Investigating effectiveness of parametric and non-parametric memories},
  author={Mallen, Alex and Asai, Akari and Zhong, Victor and Das, Rajarshi and Khashabi, Daniel and Hajishirzi, Hannaneh},
  booktitle={Proceedings of the 61st annual meeting of the association for computational linguistics (volume 1: Long papers)},
  pages={9802--9822},
  year={2023}
}

@inproceedings{yang2018hotpotqa,
  title={HotpotQA: A dataset for diverse, explainable multi-hop question answering},
  author={Yang, Zhilin and Qi, Peng and Zhang, Saizheng and Bengio, Yoshua and Cohen, William and Salakhutdinov, Ruslan and Manning, Christopher D},
  booktitle={Proceedings of the 2018 conference on empirical methods in natural language processing},
  pages={2369--2380},
  year={2018}
}

@inproceedings{ho2020constructing,
  title={Constructing a multi-hop qa dataset for comprehensive evaluation of reasoning steps},
  author={Ho, Xanh and Nguyen, Anh-Khoa Duong and Sugawara, Saku and Aizawa, Akiko},
  booktitle={Proceedings of the 28th International Conference on Computational Linguistics},
  pages={6609--6625},
  year={2020}
}

@article{trivedi2022musique,
  title={MuSiQue: Multihop Questions via Single-hop Question Composition},
  author={Trivedi, Harsh and Balasubramanian, Niranjan and Khot, Tushar and Sabharwal, Ashish},
  journal={Transactions of the Association for Computational Linguistics},
  volume={10},
  pages={539--554},
  year={2022},
  publisher={MIT Press One Broadway, 12th Floor, Cambridge, Massachusetts 02142, USA~…}
}

@inproceedings{press2023measuring,
  title={Measuring and narrowing the compositionality gap in language models},
  author={Press, Ofir and Zhang, Muru and Min, Sewon and Schmidt, Ludwig and Smith, Noah A and Lewis, Mike},
  booktitle={Findings of the Association for Computational Linguistics: EMNLP 2023},
  pages={5687--5711},
  year={2023}
}

@article{yang2025qwen3,
  title={Qwen3 technical report},
  author={Yang, An and Li, Anfeng and Yang, Baosong and Zhang, Beichen and Hui, Binyuan and Zheng, Bo and Yu, Bowen and Gao, Chang and Huang, Chengen and Lv, Chenxu and others},
  journal={arXiv preprint arXiv:2505.09388},
  year={2025}
}

@article{jin2025search,
  title={Search-r1: Training llms to reason and leverage search engines with reinforcement learning},
  author={Jin, Bowen and Zeng, Hansi and Yue, Zhenrui and Yoon, Jinsung and Arik, Sercan and Wang, Dong and Zamani, Hamed and Han, Jiawei},
  journal={arXiv preprint arXiv:2503.09516},
  year={2025}
}

@article{zhang2025web,
  title={From web search towards agentic deep research: Incentivizing search with reasoning agents},
  author={Zhang, Weizhi and Li, Yangning and Bei, Yuanchen and Luo, Junyu and Wan, Guancheng and Yang, Liangwei and Xie, Chenxuan and Yang, Yuyao and Huang, Wei-Chieh and Miao, Chunyu and others},
  journal={arXiv preprint arXiv:2506.18959},
  year={2025}
}

@inproceedings{zheng2025deepresearcher,
  title={Deepresearcher: Scaling deep research via reinforcement learning in real-world environments},
  author={Zheng, Yuxiang and Fu, Dayuan and Hu, Xiangkun and Cai, Xiaojie and Ye, Lyumanshan and Lu, Pengrui and Liu, Pengfei},
  booktitle={Proceedings of the 2025 Conference on Empirical Methods in Natural Language Processing},
  pages={414--431},
  year={2025}
}

@article{wang2026openclaw,
  title={Openclaw-rl: Train any agent simply by talking},
  author={Wang, Yinjie and Chen, Xuyang and Jin, Xiaolong and Wang, Mengdi and Yang, Ling},
  journal={arXiv preprint arXiv:2603.10165},
  year={2026}
}

@article{dong2025agentic,
  title={Agentic reinforced policy optimization},
  author={Dong, Guanting and Mao, Hangyu and Ma, Kai and Bao, Licheng and Chen, Yifei and Wang, Zhongyuan and Chen, Zhongxia and Du, Jiazhen and Wang, Huiyang and Zhang, Fuzheng and others},
  journal={arXiv preprint arXiv:2507.19849},
  year={2025}
}

@article{lu2026self,
  title={Self-distilled agentic reinforcement learning},
  author={Lu, Zhengxi and Yao, Zhiyuan and Han, Zhuowen and Wang, Zi-Han and Wu, Jinyang and Gu, Qi and Cai, Xunliang and Lu, Weiming and Xiao, Jun and Zhuang, Yueting and others},
  journal={arXiv preprint arXiv:2605.15155},
  year={2026}
}

@inproceedings{agarwal2024policy,
  title={On-policy distillation of language models: Learning from self-generated mistakes},
  author={Agarwal, Rishabh and Vieillard, Nino and Zhou, Yongchao and Stanczyk, Piotr and Ramos Garea, Sabela and Geist, Matthieu and Bachem, Olivier},
  booktitle={International Conference on Learning Representations},
  volume={2024},
  pages={21246--21263},
  year={2024}
}

@article{li2026rethinking,
  title={Rethinking on-policy distillation of large language models: Phenomenology, mechanism, and recipe},
  author={Li, Yaxuan and Zuo, Yuxin and He, Bingxiang and Zhang, Jinqian and Xiao, Chaojun and Qian, Cheng and Yu, Tianyu and Gao, Huan-ang and Yang, Wenkai and Liu, Zhiyuan and others},
  journal={arXiv preprint arXiv:2604.13016},
  year={2026}
}

@article{gudibande2023false,
  title={The false promise of imitating proprietary llms},
  author={Gudibande, Arnav and Wallace, Eric and Snell, Charlie and Geng, Xinyang and Liu, Hao and Abbeel, Pieter and Levine, Sergey and Song, Dawn},
  journal={arXiv preprint arXiv:2305.15717},
  year={2023}
}

@article{zhao2026self,
  title={Self-Distilled Reasoner: On-Policy Self-Distillation for Large Language Models},
  author={Zhao, Siyan and Xie, Zhihui and Liu, Mengchen and Huang, Jing and Pang, Guan and Chen, Feiyu and Grover, Aditya},
  journal={arXiv preprint arXiv:2601.18734},
  year={2026}
}

@article{shao2024deepseekmath,
  title={Deepseekmath: Pushing the limits of mathematical reasoning in open language models, 2024},
  author={Shao, Zhihong and Wang, Peiyi and Zhu, Qihao and Xu, Runxin and Song, Junxiao and Bi, Xiao and Zhang, Haowei and Zhang, Mingchuan and Li, YK and Wu, Yang and others},
  journal={URL https://arxiv. org/abs/2402.03300},
  volume={2},
  number={3},
  pages={5},
  year={2024}
}

@inproceedings{liu2026amo,
  title={Amo-bench: Large language models still struggle in high school math competitions},
  author={Liu, Junlin and An, Shengnan and Zhou, Shuang and Ma, Dan and Lin, Yehao and Lv, Xinxuan and Wang, Xuanlin and Li, Xiaoyu and Wang, Ziwen and Cao, Xuezhi and others},
  booktitle={Findings of the Association for Computational Linguistics: ACL 2026},
  pages={2120--2137},
  year={2026}
}

@article{liu2026general365,
  title={General365: Benchmarking General Reasoning in Large Language Models Across Diverse and Challenging Tasks},
  author={Liu, Junlin and An, Shengnan and Zhou, Shuang and Ma, Dan and Luo, Shixiong and Xie, Ying and Zhang, Yuan and Yuan, Wenling and Zhou, Yifan and Li, Xiaoyu and others},
  journal={arXiv preprint arXiv:2604.11778},
  year={2026}
}

@article{dong2025agentic2,
  title={Agentic entropy-balanced policy optimization},
  author={Dong, Guanting and Bao, Licheng and Wang, Zhongyuan and Zhao, Kangzhi and Li, Xiaoxi and Jin, Jiajie and Yang, Jinghan and Mao, Hangyu and Zhang, Fuzheng and Gai, Kun and others},
  journal={arXiv preprint arXiv:2510.14545},
  year={2025}
}

@article{lv2026physagent,
  title={PhysAgent: Automating Physics-Based 4D Synthesis via Trajectory-Grounded Multi-Agent Feedback},
  author={Lv, Chunji and Ye, Jiaxi and Jiang, Yuchen and Lin, Rexar and Li, Changsheng},
  journal={arXiv preprint arXiv:2606.08688},
  year={2026}
}

@article{mai2025agent,
  title={Agent rl scaling law: Agent rl with spontaneous code execution for mathematical problem solving},
  author={Mai, Xinji and Xu, Haotian and Li, Zhong-Zhi and Wang, Weinong and Hu, Jian and Zhang, Yingying and Zhang, Wenqiang and others},
  journal={arXiv preprint arXiv:2505.07773},
  year={2025}
}

@article{wang2025ragen,
  title={Ragen: Understanding self-evolution in llm agents via multi-turn reinforcement learning},
  author={Wang, Zihan and Wang, Kangrui and Wang, Qineng and Zhang, Pingyue and Li, Linjie and Yang, Zhengyuan and Jin, Xing and Yu, Kefan and Nguyen, Minh Nhat and Liu, Licheng and others},
  journal={arXiv preprint arXiv:2504.20073},
  year={2025}
}

@inproceedings{qi2025webrl,
  title={Webrl: Training llm web agents via self-evolving online curriculum reinforcement learning},
  author={Qi, Zehan and Liu, Xiao and Iong, Iat Long and Lai, Hanyu and Sun, Xueqiao and Sun, Jiadai and Yang, Xinyue and Yang, Yu and Yao, Shuntian and Xu, Wei and others},
  booktitle={International Conference on Learning Representations},
  volume={2025},
  pages={79791--79821},
  year={2025}
}

@inproceedings{hsieh2023distilling,
  title={Distilling step-by-step! outperforming larger language models with less training data and smaller model sizes},
  author={Hsieh, Cheng-Yu and Li, Chun-Liang and Yeh, Chih-Kuan and Nakhost, Hootan and Fujii, Yasuhisa and Ratner, Alex and Krishna, Ranjay and Lee, Chen-Yu and Pfister, Tomas},
  booktitle={Findings of the Association for Computational Linguistics: ACL 2023},
  pages={8003--8017},
  year={2023}
}

@article{mukherjee2023orca,
  title={Orca: Progressive learning from complex explanation traces of gpt-4},
  author={Mukherjee, Subhabrata and Mitra, Arindam and Jawahar, Ganesh and Agarwal, Sahaj and Palangi, Hamid and Awadallah, Ahmed},
  journal={arXiv preprint arXiv:2306.02707},
  year={2023}
}

@inproceedings{xu2025magpie,
  title={Magpie: Alignment data synthesis from scratch by prompting aligned llms with nothing},
  author={Xu, Zhangchen and Jiang, Fengqing and Niu, Luyao and Deng, Yuntian and Poovendran, Radha and Choi, Yejin and Lin, Bill Yuchen},
  booktitle={International Conference on Learning Representations},
  volume={2025},
  pages={76346--76382},
  year={2025}
}

@article{hinton2015distilling,
  title={Distilling the knowledge in a neural network},
  author={Hinton, Geoffrey and Vinyals, Oriol and Dean, Jeff},
  journal={arXiv preprint arXiv:1503.02531},
  year={2015}
}

@article{wu2023autogen,
  title={Autogen: Enabling next-gen llm applications via multi-agent conversation},
  author={Wu, Qingyun and Bansal, Gagan and Zhang, Jieyu and Wu, Yiran and Li, Beibin and Zhu, Erkang and Jiang, Li and Zhang, Xiaoyun and Zhang, Shaokun and Liu, Jiale and others},
  journal={arXiv preprint arXiv:2308.08155},
  year={2023}
}

@inproceedings{hong2024metagpt,
  title={MetaGPT: Meta programming for a multi-agent collaborative framework},
  author={Hong, Sirui and Zhuge, Mingchen and Chen, Jonathan and Zheng, Xiawu and Cheng, Yuheng and Wang, Jinlin and Zhang, Ceyao and Yau, Steven and Lin, Zijuan and Zhou, Liyang and others},
  booktitle={International Conference on Learning Representations},
  volume={2024},
  pages={23247--23275},
  year={2024}
}

@article{talebirad2023multi,
  title={Multi-agent collaboration: Harnessing the power of intelligent llm agents},
  author={Talebirad, Yashar and Nadiri, Amirhossein},
  journal={arXiv preprint arXiv:2306.03314},
  year={2023}
}

@article{du2023improving,
  title={Improving factuality and reasoning in language models through multiagent debate},
  author={Du, Yilun and Li, Shuang and Torralba, Antonio and Tenenbaum, Joshua B and Mordatch, Igor},
  journal={arXiv preprint arXiv:2305.14325},
  year={2023}
}

@article{lei2024macm,
  title={Macm: Utilizing a multi-agent system for condition mining in solving complex mathematical problems},
  author={Lei, Bin and Zhang, Yi and Zuo, Shan and Payani, Ali and Ding, Caiwen},
  journal={Advances in Neural Information Processing Systems},
  volume={37},
  pages={53418--53437},
  year={2024}
}

@article{wang2026mad,
  title={Mad-opd: Breaking the ceiling in on-policy distillation via multi-agent debate},
  author={Wang, Jianze and Liu, Ying and Chen, Jinlong and Hu, Xuchun and Zhang, Qilong and Cao, Yu and Wang, Jun and Yang, Hua and Xie, Yong and Chen, Qianglong},
  journal={arXiv preprint arXiv:2605.01347},
  year={2026}
}

@article{jiang2026beyond,
  title={Beyond Relevance-Centric Retrieval: Rubric-Oriented Document Set Selection and Ranking},
  author={Jiang, Kailin and Liu, Lei and Xi, Jian and Xu, Hui and Liu, Junlin and Fu, Baochen and Ren, Shaoqing and Li, Bin and Lu, Yu and Shi, Haibo and others},
  journal={arXiv preprint arXiv:2607.19747},
  year={2026}
}

@article{jiang2025kore,
  title={KORE: Enhancing Knowledge Injection for Large Multimodal Models via Knowledge-Oriented Augmentations and Constraints},
  author={Jiang, Kailin and Jiang, Hongbo and Jiang, Ning and Gao, Zhi and Bi, Jinhe and Ren, Yuchen and Li, Bin and Du, Yuntao and Liu, Lei and Li, Qing},
  journal={arXiv preprint arXiv:2510.19316},
  year={2025}
}

@inproceedings{jiang2026mined,
  title={Mined: Probing and updating with multimodal time-sensitive knowledge for large multimodal models},
  author={Jiang, Kailin and Jiang, Ning and Du, Yuntao and Ren, Yuchen and Li, Yuchen and Gao, Yifan and Bi, Jinhe and Ma, Yunpu and Li, Bin and Liu, Lei and others},
  booktitle={Findings of the Association for Computational Linguistics: ACL 2026},
  pages={13766--13795},
  year={2026}
}

@inproceedings{jiang2026large,
  title={When large multimodal models confront evolving knowledge: Challenges and explorations},
  author={Jiang, Kailin and Du, Yuntao and Ding, Yukai and Ren, Yuchen and Jiang, Ning and Gao, Zhi and Zheng, Zilong and Liu, Lei and Li, Bin and Li, Qing},
  booktitle={The Fourteenth International Conference on Learning Representations},
  year={2026}
}

@inproceedings{jia2026benchmarking,
  title={Benchmarking multimodal knowledge conflict for large multimodal models},
  author={Jia, Yifan and Du, Yuntao and Jiang, Kailin and Liang, Yuyang and Ren, Qihan and Xin, Yi and Yang, Rui and Feng, Fenze and Chen, MingCai and Lu, Hengyang and others},
  booktitle={Proceedings of the AAAI Conference on Artificial Intelligence},
  volume={40},
  number={27},
  pages={22283--22291},
  year={2026}
}

@inproceedings{jiang2025mmke,
  title={Mmke-bench: A multimodal editing benchmark for diverse visual knowledge},
  author={Jiang, Kailin and Gao, Zhi and Shi, Chenrui and Zheng, Zilong and Qi, Siyuan and Li, Qing and others},
  booktitle={International Conference on Learning Representations},
  volume={2025},
  pages={526--555},
  year={2025}
}

@inproceedings{qi2025context,
  title={In-context editing: Learning knowledge from self-induced distributions},
  author={Qi, Siyuan and Yang, Bangcheng and Jiang, Kailin and Wang, Xiaobo and Li, Jiaqi and Zhong, Yifan and Yang, Yaodong and Zheng, Zilong},
  booktitle={International Conference on Learning Representations},
  volume={2025},
  pages={77563--77585},
  year={2025}
}

@inproceedings{Fu2026CanML,
  title={Can Multimodal Large Language Models Understand OCT?},
  author={Baochen Fu and Wenzhi Deng and Baihao Jin and Yang Li and Zihan Nie and Kailin Jiang and Yuntao Du and Weiye Song},
  year={2026},
}

@article{Fu2026MMKUBenchAM,
  title={MMKU-Bench: A Multimodal Update Benchmark for Diverse Visual Knowledge},
  author={Baochen Fu and Yuntao Du and Cheng-Wei Chang and Baihao Jin and Wenzhi Deng and Muhao Xu and Hongmei Yan and Weiye Song and Yi Wan},
  journal={ArXiv},
  year={2026},
  volume={abs/2603.15117},
}
\bibliographystyle{conference}

\appendix
\clearpage

\section{Structured Protocol Example}\label{app:protocol_example}

Example~\ref{protocolexmp:protocol_claude}, Example~\ref{protocolexmp:protocol_GPT} and Example~\ref{protocolexmp:protocol_gemini} show three complete structured protocols produced by our MAS pipeline with different teacher backbones (Claude-Opus-4.6, GPT-5.5 and Gemini-3.1-Pro, respectively).

\begin{figure*}[ht]
\small
\begin{protocolexmp}{Structured Protocol Example of Claude-Opus-4.6}{protocol_claude}
{}\textbf{Question:} Which English actor and director who was involved with the 1997 film \textit{The Winter Guest} also appeared in ``Die Hard''? \par
\textbf{Ground-Truth Answer:} Alan Rickman \par
\vspace{4pt}
\hrule
\vspace{6pt}
\texttt{\textbf{"task\_type":}} \texttt{\textbf{"multi\_hop"}} \par
\vspace{4pt}
\texttt{\textbf{"reasoning\_plan":}} \textcolor{plancolor}{[} \par
~~~~~\textcolor{plancolor}{1. ``Search for \textit{The Winter Guest} 1997 film to identify its director and cast.''} \par
~~~~~\textcolor{plancolor}{2. ``Search for \textit{Die Hard} 1988 film to identify its cast.''} \par
~~~~~\textcolor{plancolor}{3. ``Cross-reference the two casts to find the English actor and director common to both.''} \par
\textcolor{plancolor}{]} \par
\vspace{4pt}
\texttt{\textbf{"grounding\_facts":}} \textcolor{factcolor}{[} \par
~~~~~\textcolor{factcolor}{1. ``The Winter Guest is a 1997 film directed by Alan Rickman in his debut as a director.''} \par
~~~~~\textcolor{factcolor}{2. ``Alan Sidney Patrick Rickman (21 February 1946\,--\,14 January 2016) was an English actor and director.''} \par
~~~~~\textcolor{factcolor}{3. ``Rickman's first film role was as the German terrorist leader Hans Gruber in `Die Hard' (1988).''} \par
\textcolor{factcolor}{]} \par
\vspace{4pt}
\texttt{\textbf{"answer":}} \textcolor{answercolor}{\texttt{\textbf{"Alan Rickman"}}} \par
\texttt{\textbf{"answer\_grounded":}} \textcolor{answercolor}{\texttt{\textbf{true}}}
\end{protocolexmp}
\label{fig:protocol_example_claude}
\end{figure*}

\begin{figure*}[ht]
\small
\begin{protocolexmp}{Structured Protocol Example of GPT-5.5}{protocol_GPT}
{}\textbf{Question:} What year was the barber surgeon who served the last French monarch of the House of Valois born? \par
\textbf{Ground-Truth Answer:} 1510 \par
\vspace{4pt}
\hrule
\vspace{6pt}
\texttt{\textbf{"task\_type":}} \texttt{\textbf{"multi\_hop"}} \par
\vspace{4pt}
\texttt{\textbf{"reasoning\_plan":}} \textcolor{plancolor}{[} \par
~~~~~\textcolor{plancolor}{1. ``Search for the last French monarch of the House of Valois to identify the king's name.''} \par
~~~~~\textcolor{plancolor}{2. ``Search for that king's name + `barber surgeon' to find the person who served him.''} \par
~~~~~\textcolor{plancolor}{3. ``Look up the barber surgeon's Wikipedia entry to verify the birth year.''} \par
\textcolor{plancolor}{]} \par
\vspace{4pt}
\texttt{\textbf{"grounding\_facts":}} \textcolor{factcolor}{[} \par
~~~~~\textcolor{factcolor}{1. ``In 1589, at the death of Henry III of France, the House of Valois became extinct in the male line.''} \par
~~~~~\textcolor{factcolor}{2. ``Ambroise Par\'{e} (c.\,1510\,--\,20 December 1590) was a French barber surgeon who served in that role for kings Henry II, Francis II, Charles IX and Henry III.''} \par
~~~~~\textcolor{factcolor}{3. ``Par\'{e} was born in 1510 in Bourg-Hersent in northwestern France.''} \par
\textcolor{factcolor}{]} \par
\vspace{4pt}
\texttt{\textbf{"answer":}} \textcolor{answercolor}{\texttt{\textbf{"1510"}}} \par
\texttt{\textbf{"answer\_grounded":}} \textcolor{answercolor}{\texttt{\textbf{true}}}
\end{protocolexmp}
\label{fig:protocol_example}
\end{figure*}

\begin{figure*}[t]
\small
\begin{protocolexmp}{Structured Protocol Example of Gemini-3.1-Pro}{protocol_gemini}
{}\textbf{Question:} \textit{The band famous for the single Waterfalls, had a singer with the nickname Left Eye, who was born on what day?} \par
\textbf{Ground-Truth Answer:} May 27, 1971 \par
\vspace{4pt}
\hrule
\vspace{6pt}
\texttt{\textbf{"task\_type":}} \texttt{\textbf{"multi\_hop"}} \par
\vspace{4pt}
\texttt{\textbf{"reasoning\_plan":}} \textcolor{plancolor}{\textbf{[}} \par
~~~~~\textcolor{plancolor}{1. ``Search for the band famous for the single Waterfalls to identify the musical group''} \par
~~~~~\textcolor{plancolor}{2. ``Search for the members of the identified band to find the singer who goes by the nickname Left Eye''} \par
~~~~~\textcolor{plancolor}{3. ``Search for the biographical details of the identified singer to find their exact date of birth''} \par
\textcolor{plancolor}{\textbf{]}} \par
\vspace{4pt}
\texttt{\textbf{"grounding\_facts":}} \textcolor{factcolor}{\textbf{[}} \par
~~~~~\textcolor{factcolor}{1. ``\,`Waterfalls' is a song by American recording group TLC.''} \par
~~~~~\textcolor{factcolor}{2. ``\,`Lisa Nicole Lopes (May 27, 1971 -- April 25, 2002), better known by her stage name Left Eye, was an American hip hop singer, rapper, songwriter, and producer.'} \par
~~~~~\textcolor{factcolor}{3. ``\,`Lopes was best known as one-third of the R\&B girl group TLC'} \par
\textcolor{factcolor}{\textbf{]}} \par
\vspace{4pt}
\texttt{\textbf{"answer":}} \textcolor{answercolor}{\texttt{\textbf{"May 27, 1971"}}} \par
\texttt{\textbf{"answer\_grounded":}} \textcolor{answercolor}{\texttt{\textbf{true}}}
\end{protocolexmp}
\label{fig:protocol_example_gemini}
\end{figure*}

\section{Training Dynamics}\label{app:training}
\begin{figure}[htbp]
      \centering
      \setlength{\tabcolsep}{1pt}
      \newlength{\imgheight}
      \settoheight{\imgheight}{\includegraphics[width=0.32\textwidth]{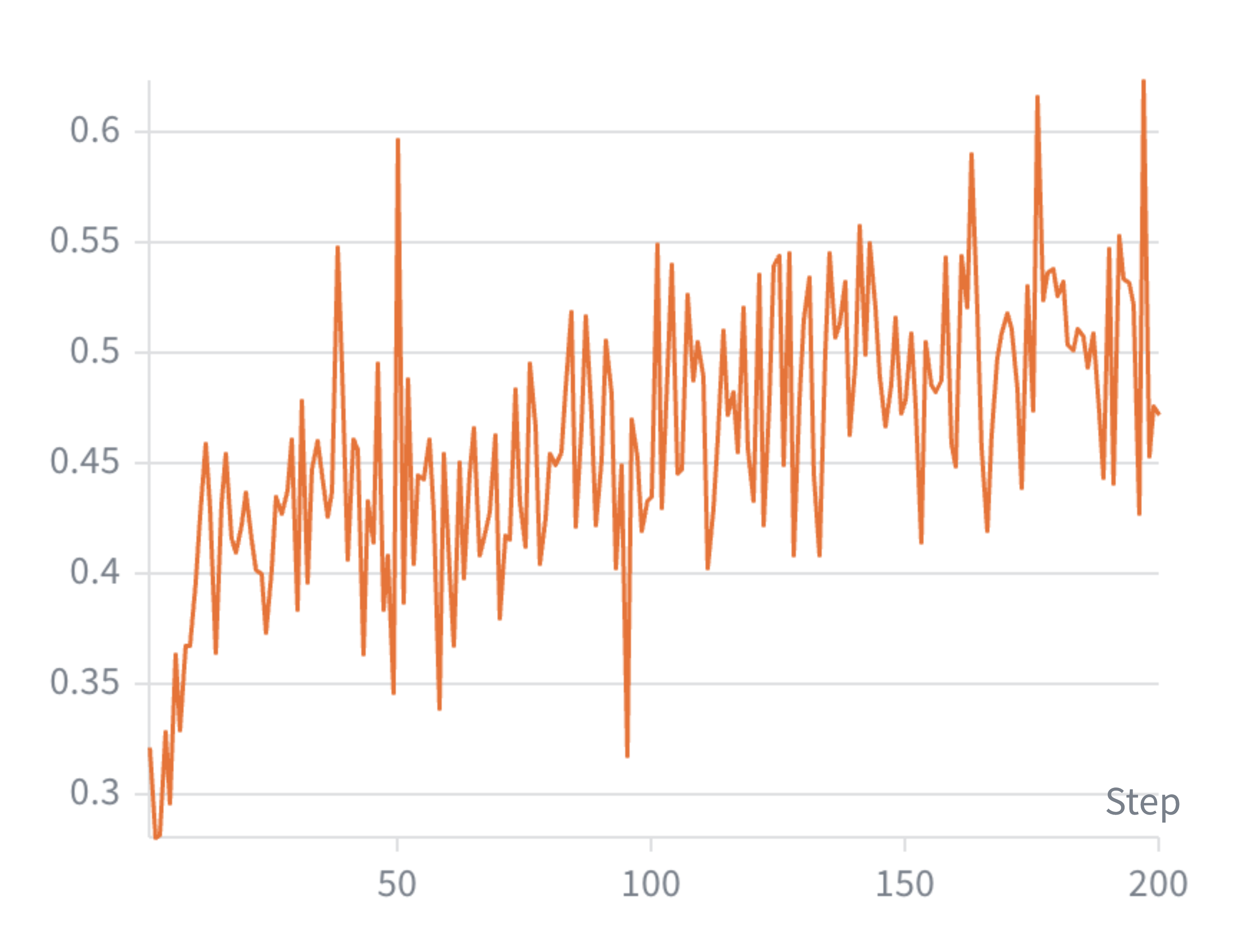}}
      \begin{tabular}{c ccc}
        & $\lambda=0.01$ & $\lambda=0.05$ & $\lambda=0.1$ \\[-2pt]
        \raisebox{0.4\imgheight}{\rotatebox{90}{\small Claude}} &
        \includegraphics[width=0.32\textwidth]{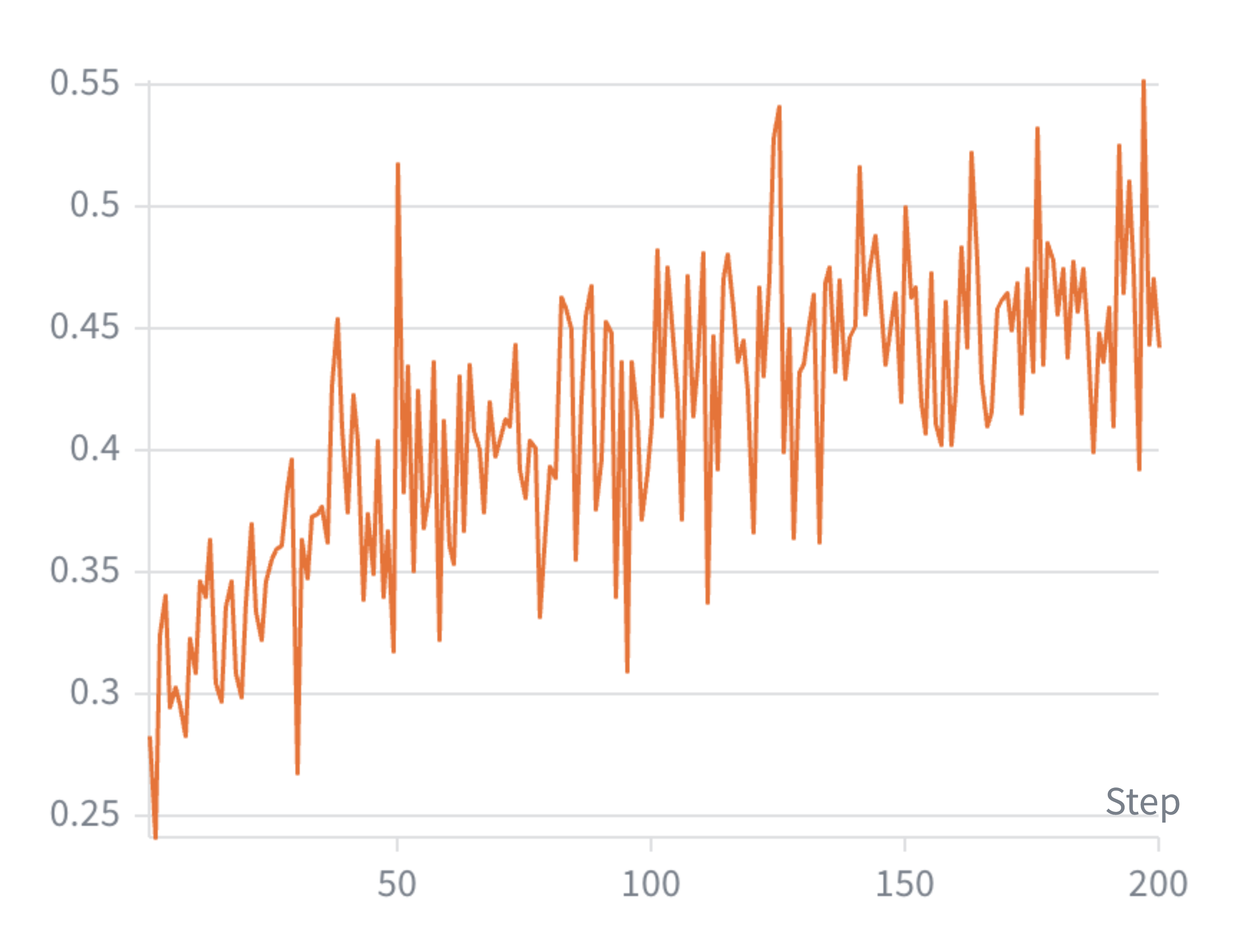} &
        \includegraphics[width=0.32\textwidth]{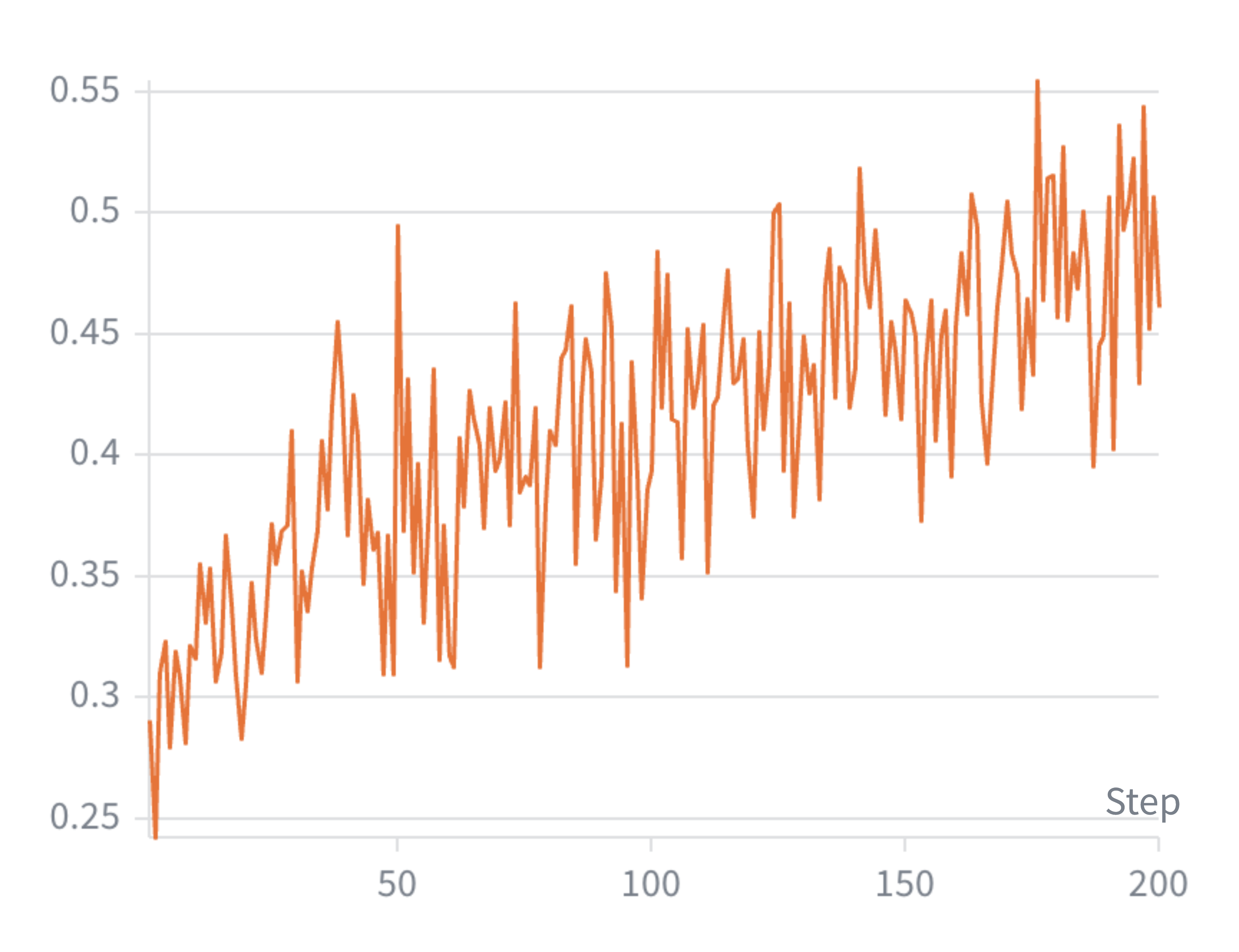} &
        \includegraphics[width=0.32\textwidth]{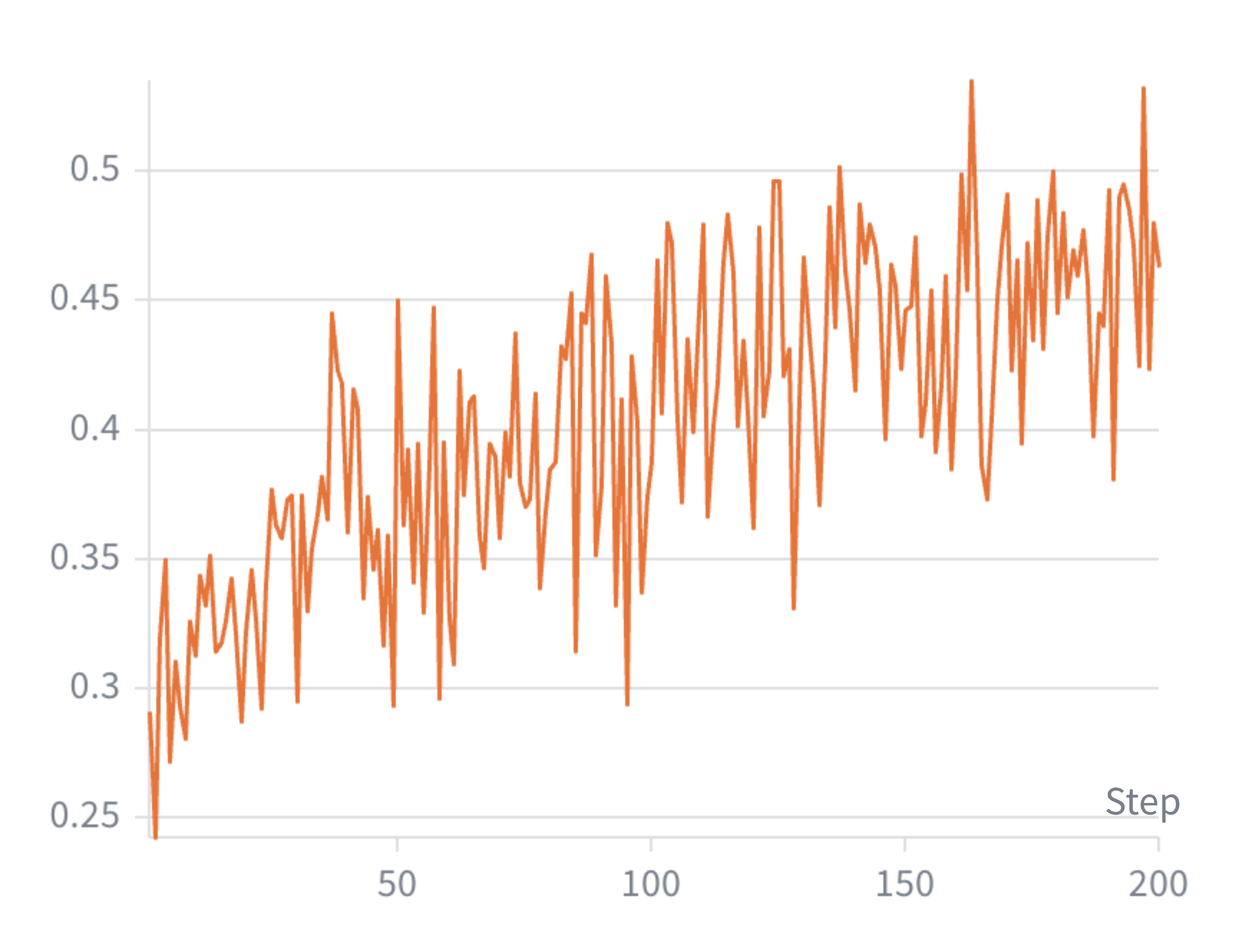} \\[-2pt]
        \raisebox{0.45\imgheight}{\rotatebox{90}{\small GPT}} &
        \includegraphics[width=0.32\textwidth]{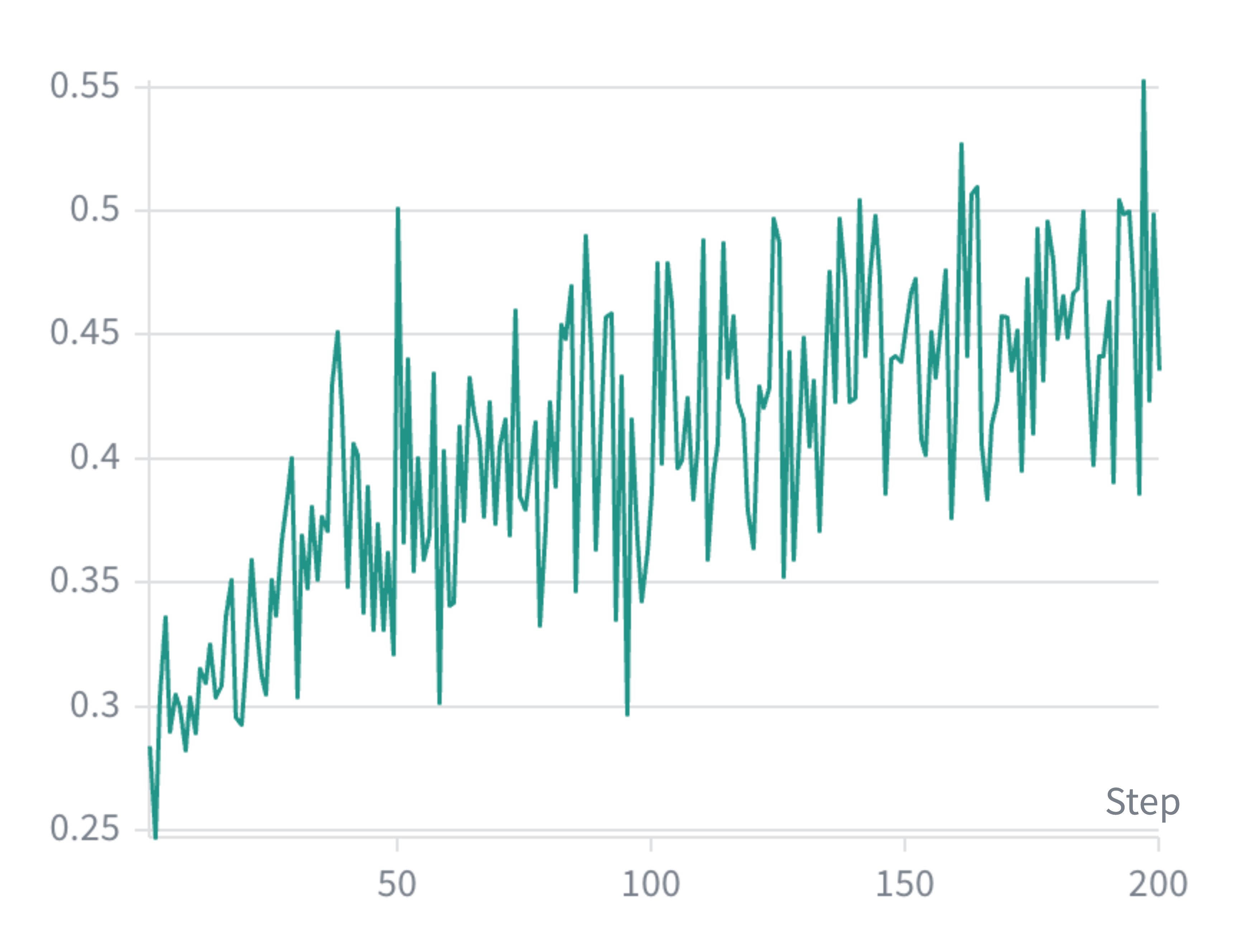} &
        \includegraphics[width=0.32\textwidth]{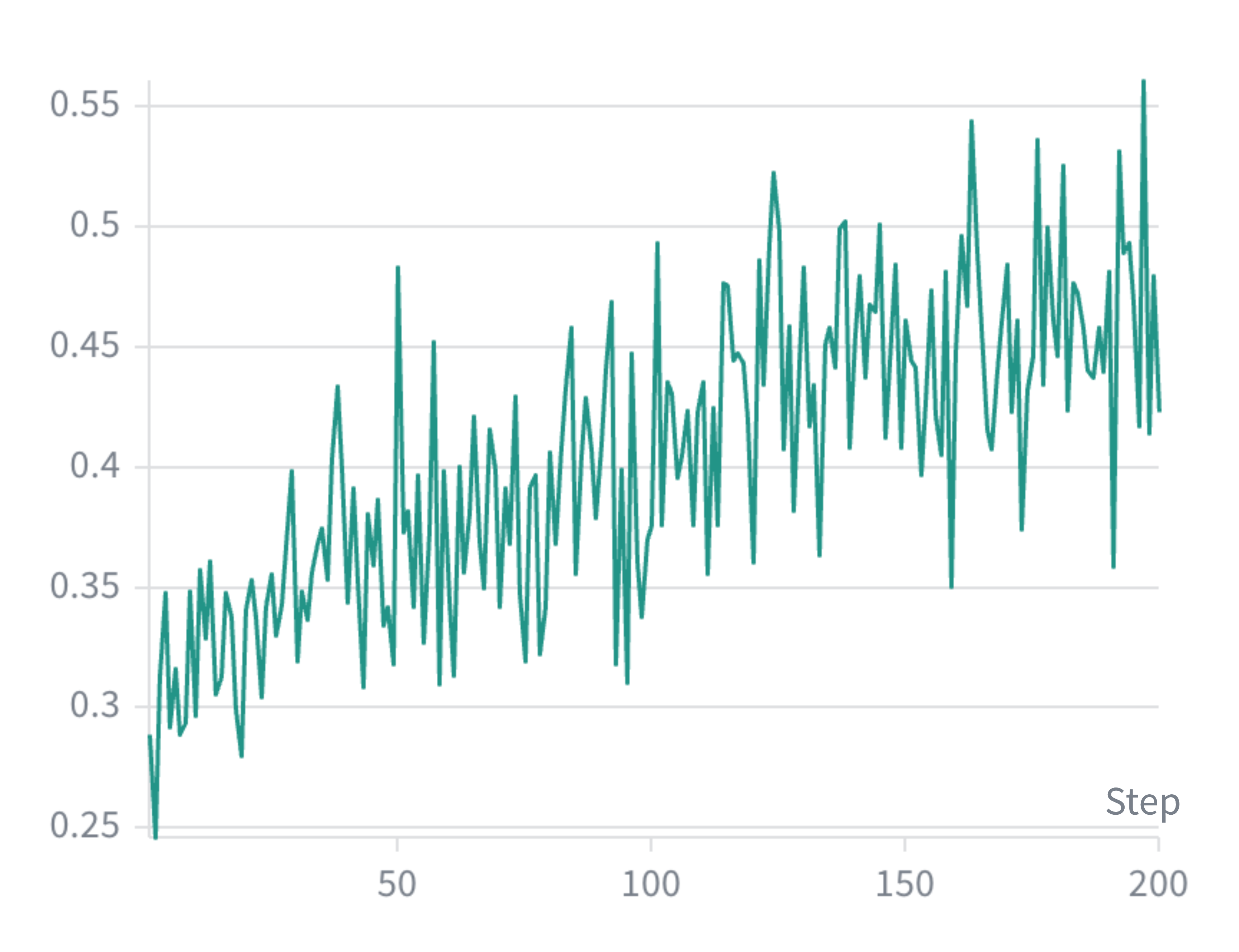} &
        \includegraphics[width=0.32\textwidth]{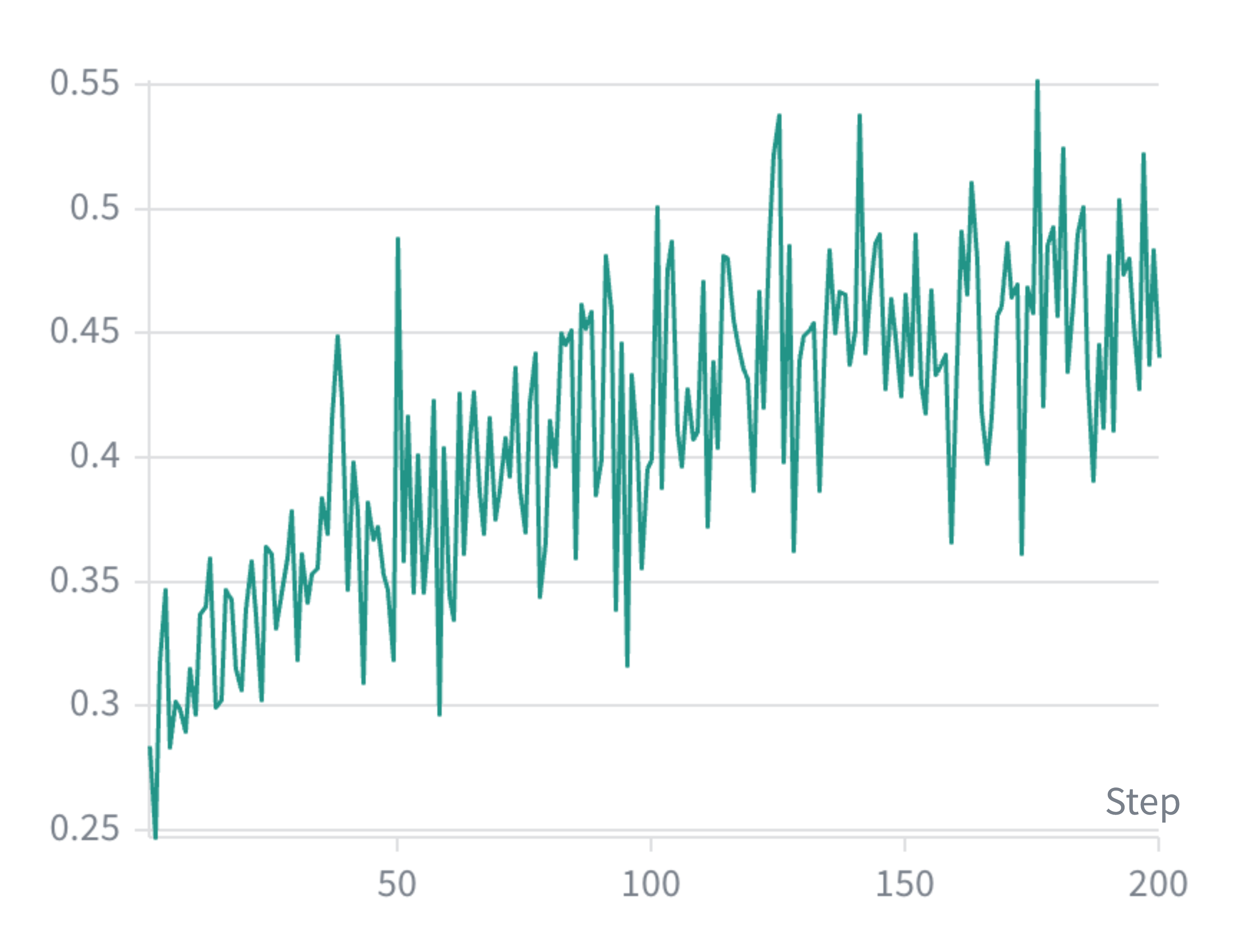} \\[-2pt]
        \raisebox{0.4\imgheight}{\rotatebox{90}{\small Gemini}} &
        \includegraphics[width=0.32\textwidth]{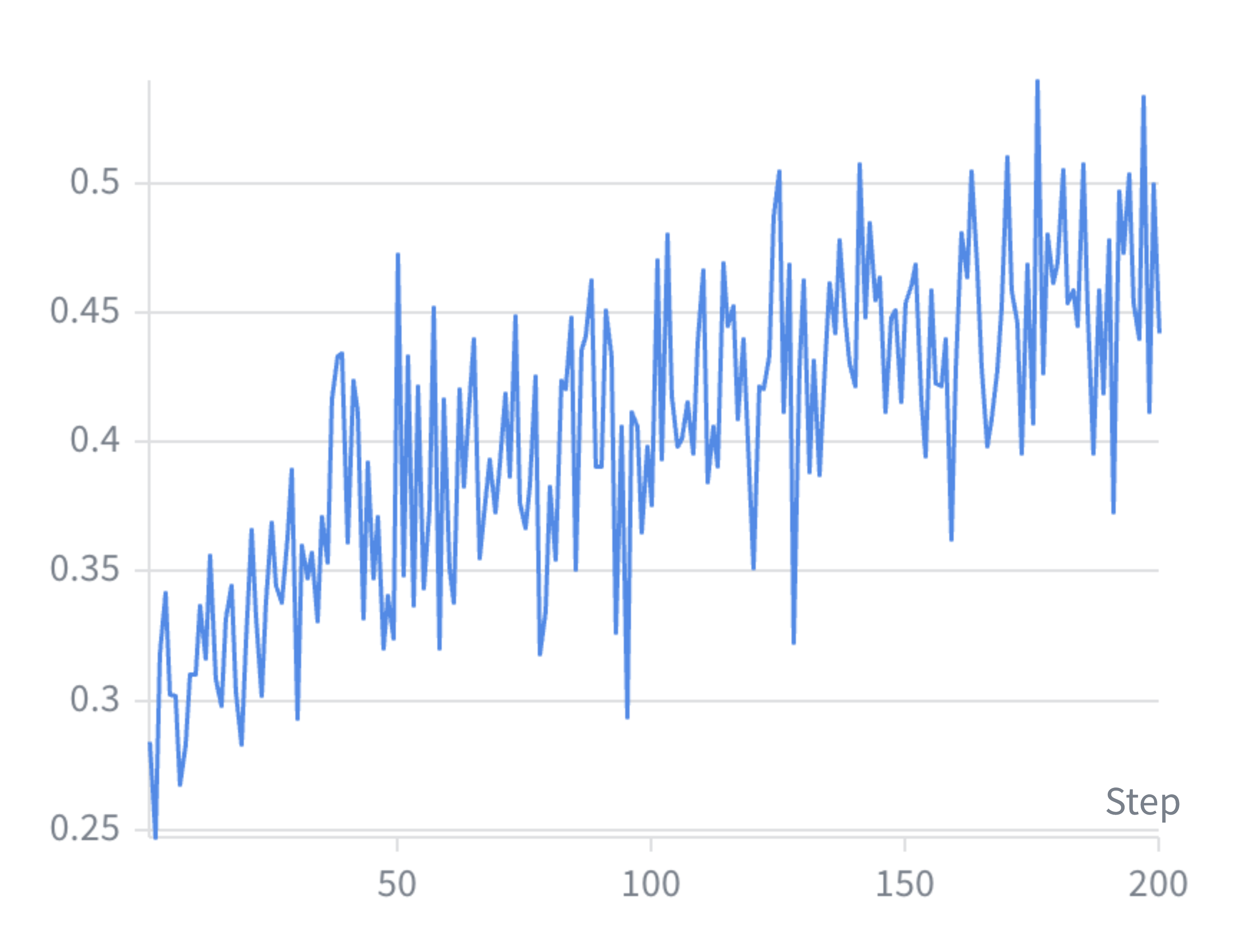} &
        \includegraphics[width=0.32\textwidth]{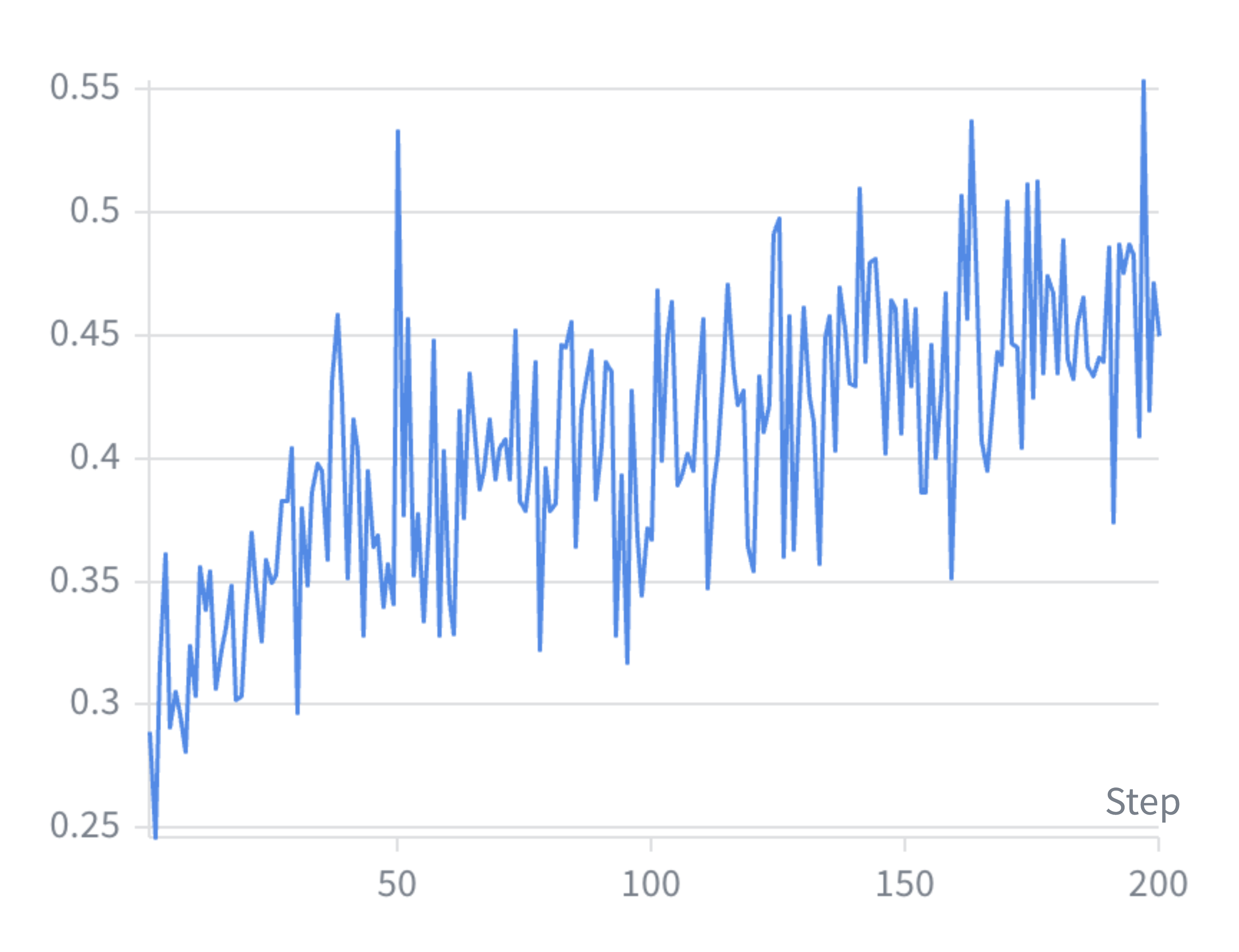} &
        \includegraphics[width=0.32\textwidth]{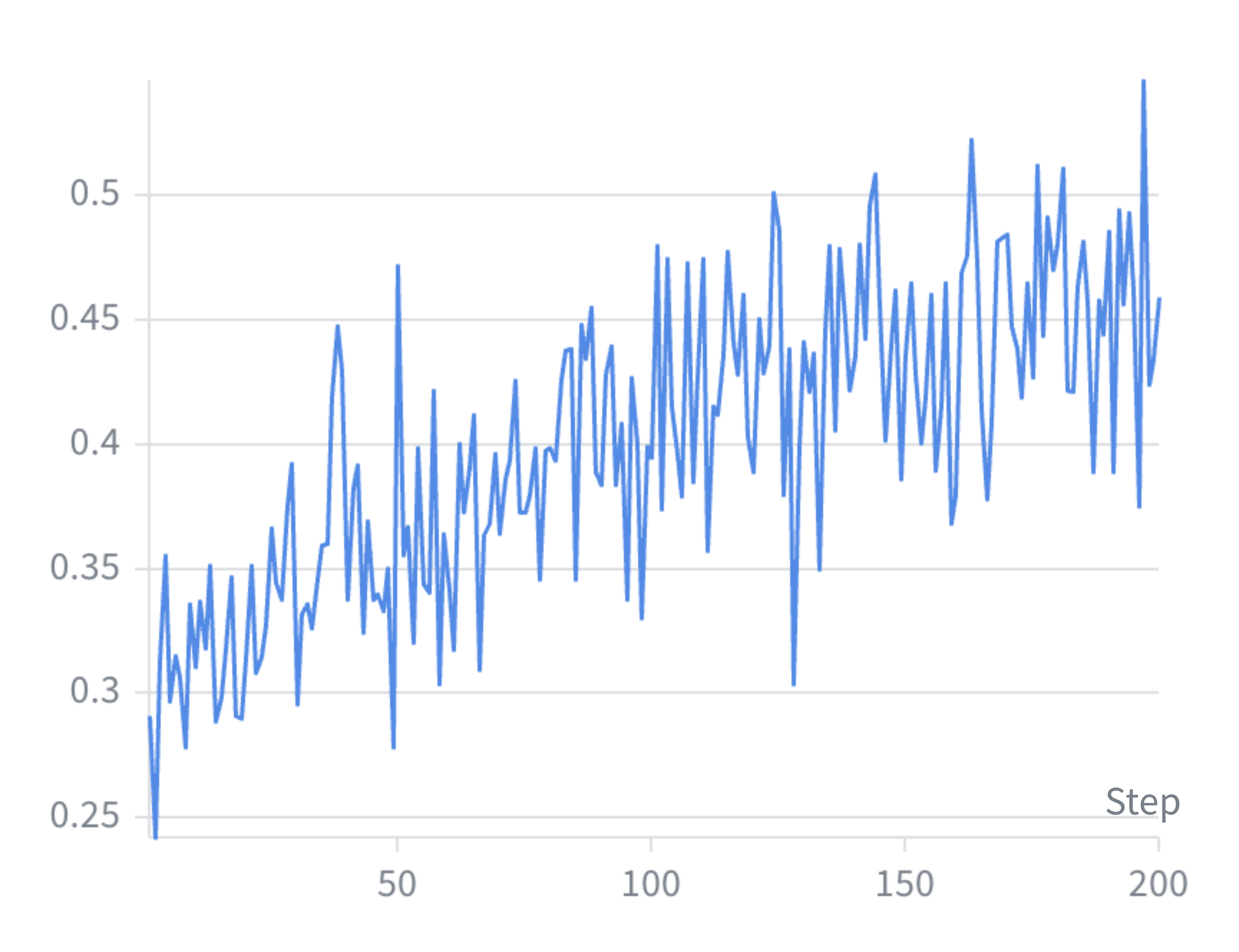} \\
      \end{tabular}
      \caption{\textbf{Training success rate} of \textbf{Qwen3-1.7B} under varying distillation weight $\lambda_{\text{OPSD}}$ with different proprietary teacher models.}
      \label{fig:training_1.7b}
  \end{figure}

\begin{figure}[htbp]
      \centering
      \setlength{\tabcolsep}{1pt}
      \settoheight{\imgheight}{\includegraphics[width=0.32\textwidth]{figures/Claude_4b_conf001.png}}
      \begin{tabular}{c ccc}
        & $\lambda=0.01$ & $\lambda=0.05$ & $\lambda=0.1$ \\[-2pt]
        \raisebox{0.4\imgheight}{\rotatebox{90}{\small Claude}} &
        \includegraphics[width=0.32\textwidth]{figures/Claude_4b_conf001.png} &
        \includegraphics[width=0.32\textwidth]{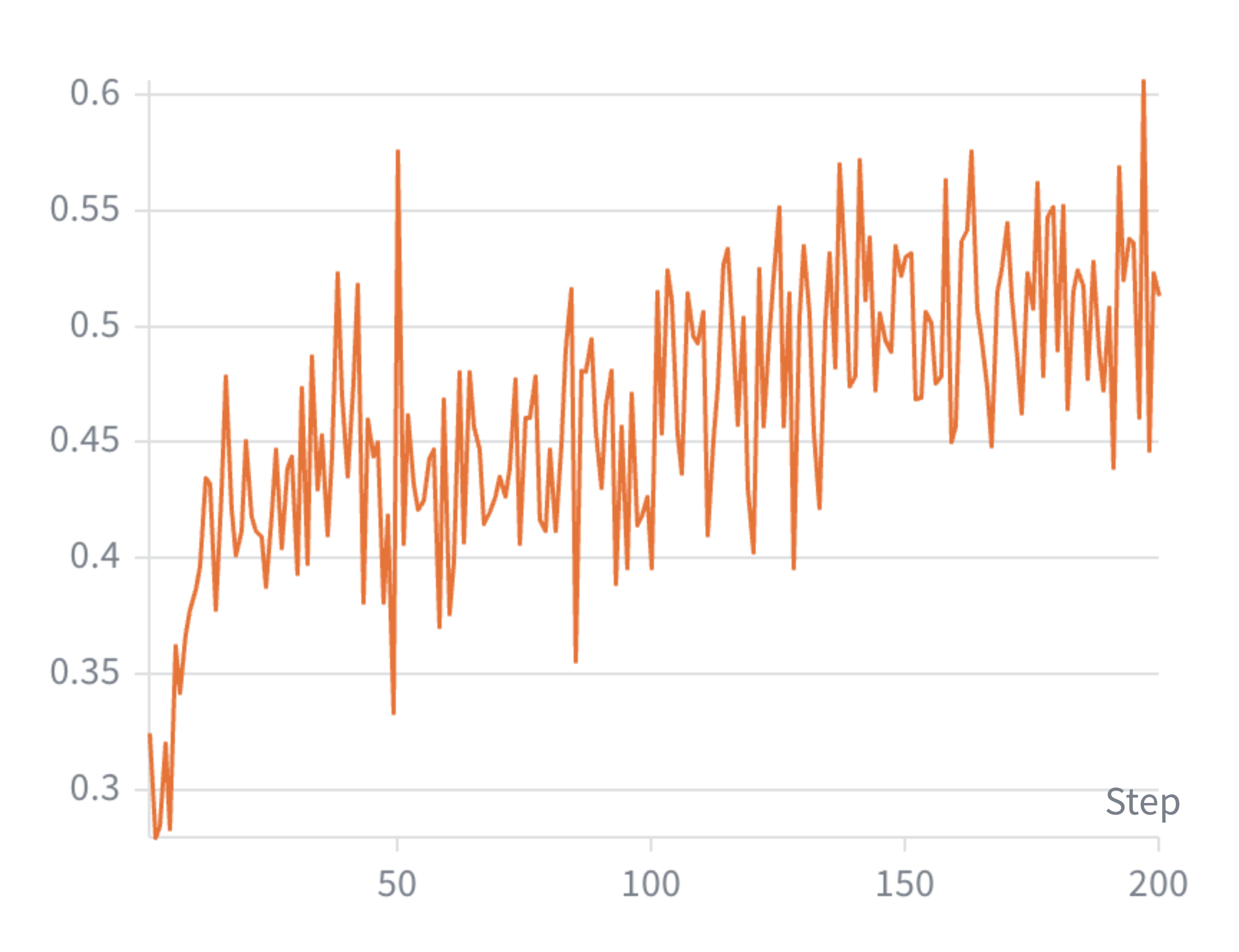} &
        \includegraphics[width=0.32\textwidth]{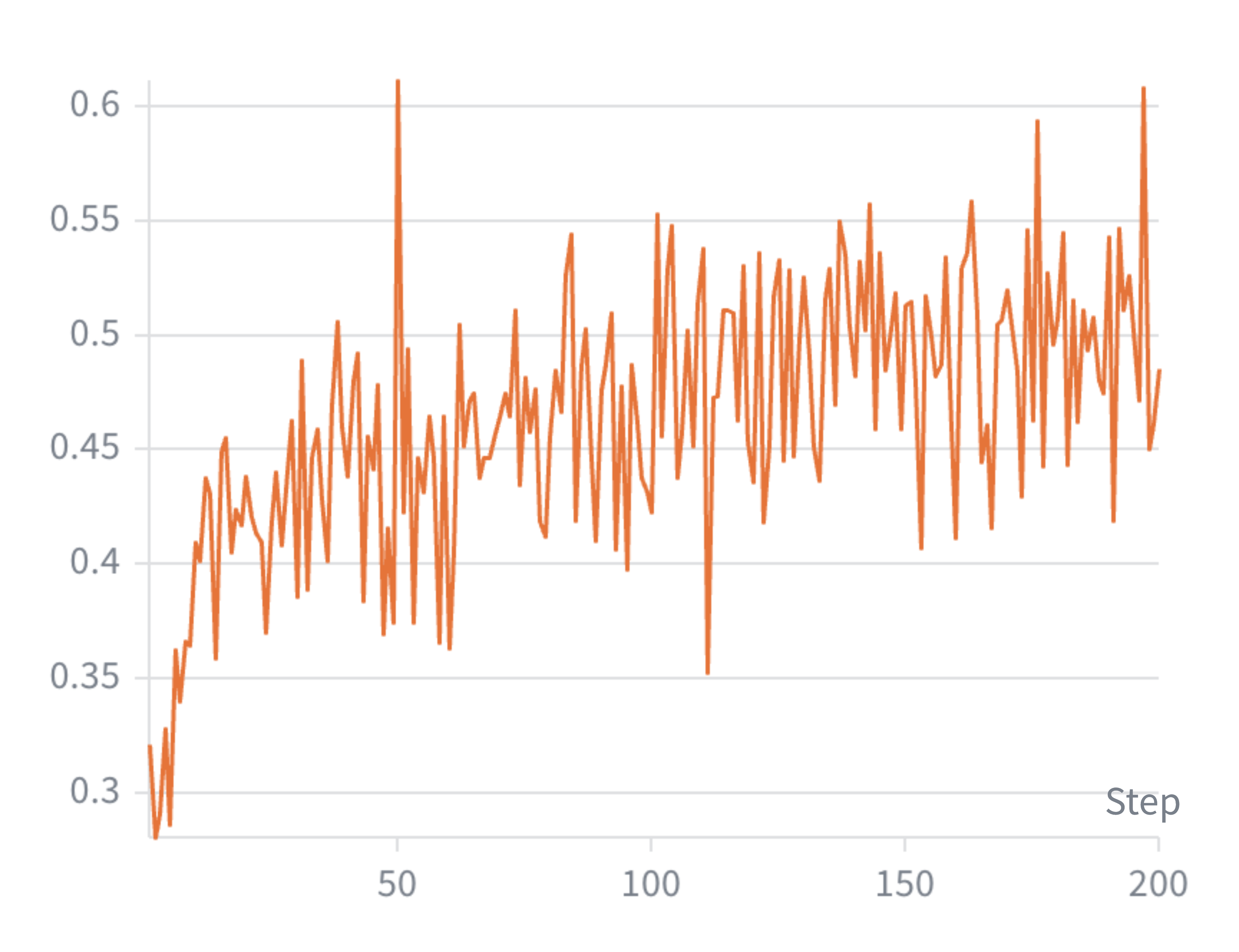} \\[-2pt]
        \raisebox{0.45\imgheight}{\rotatebox{90}{\small GPT}} &
        \includegraphics[width=0.32\textwidth]{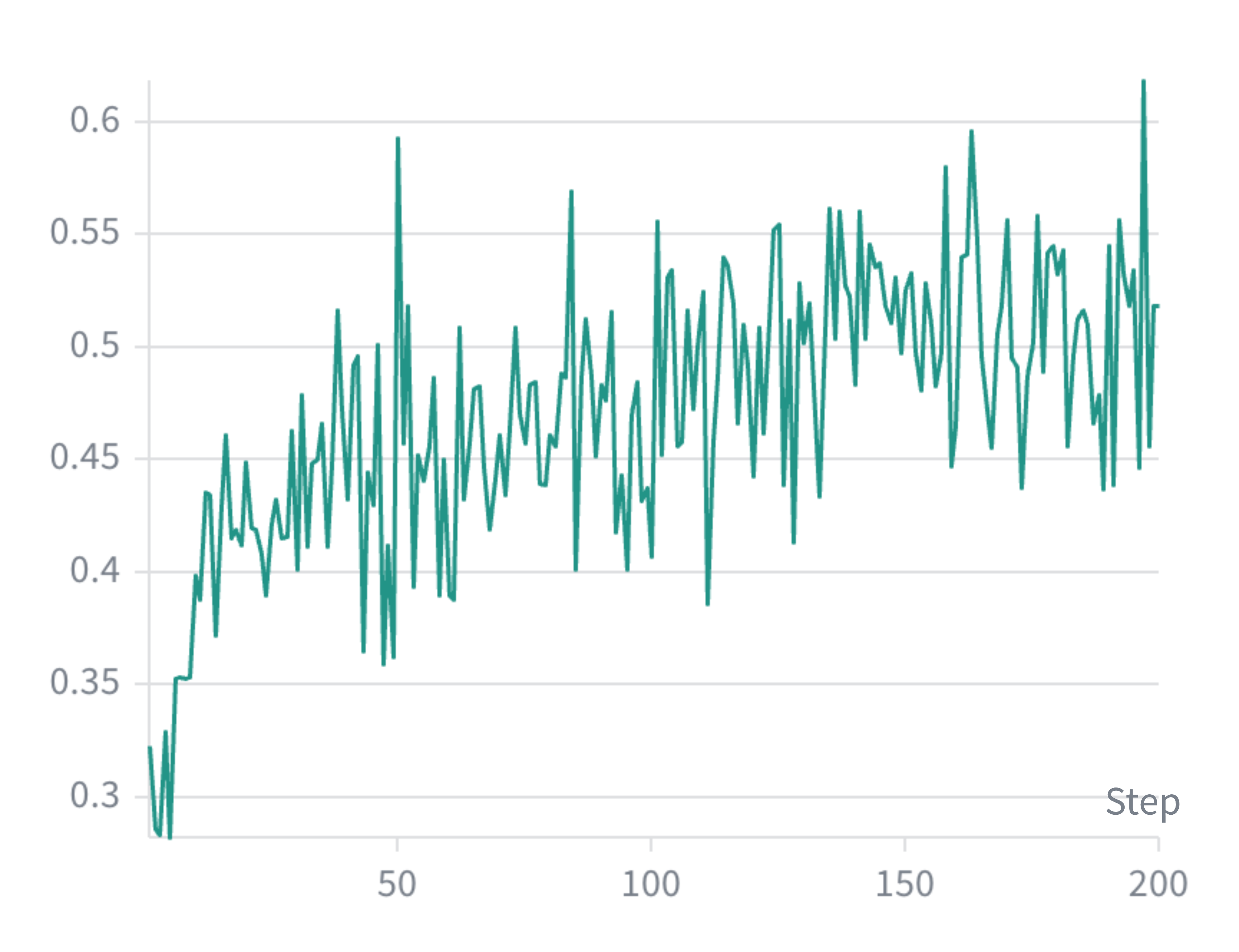} &
        \includegraphics[width=0.32\textwidth]{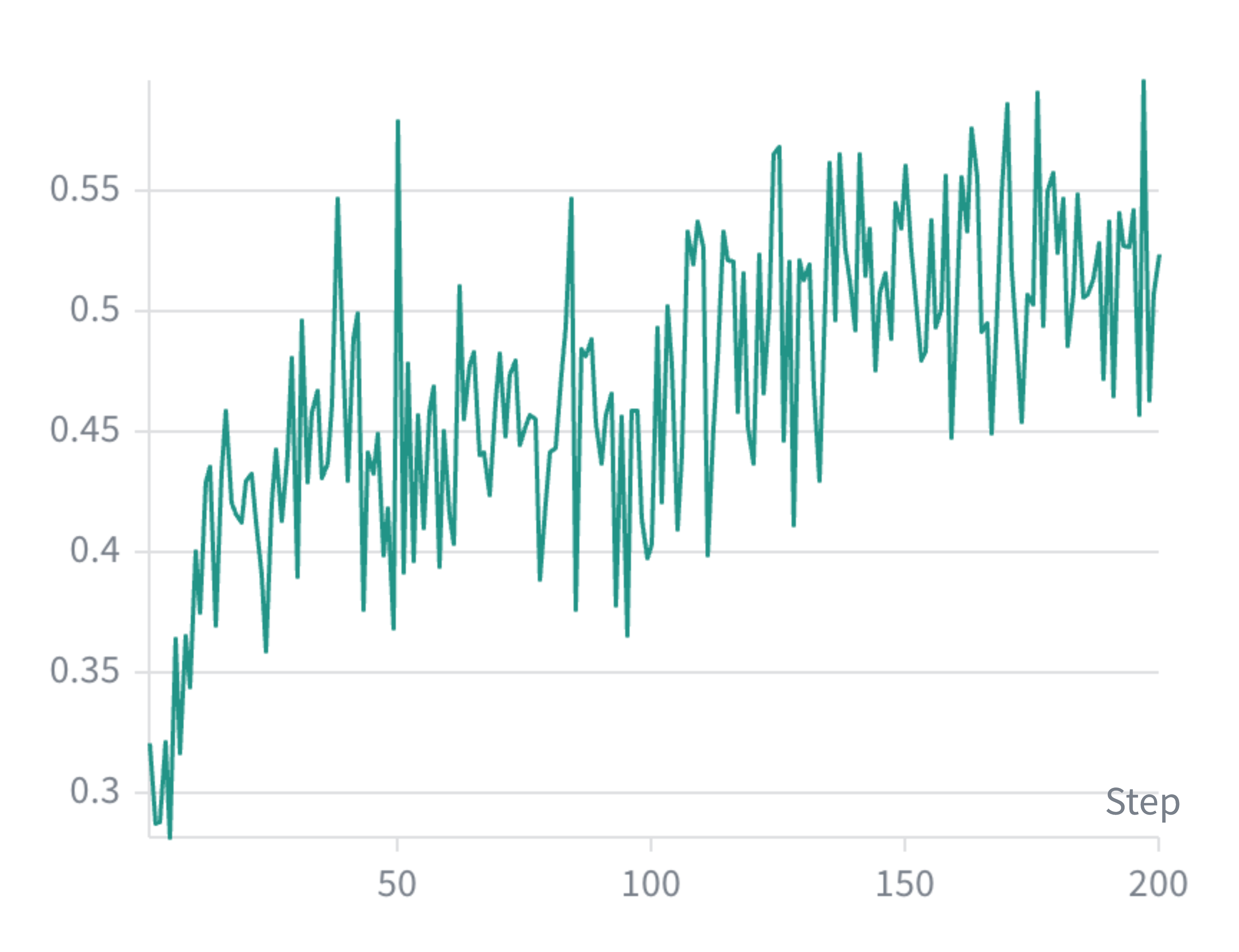} &
        \includegraphics[width=0.32\textwidth]{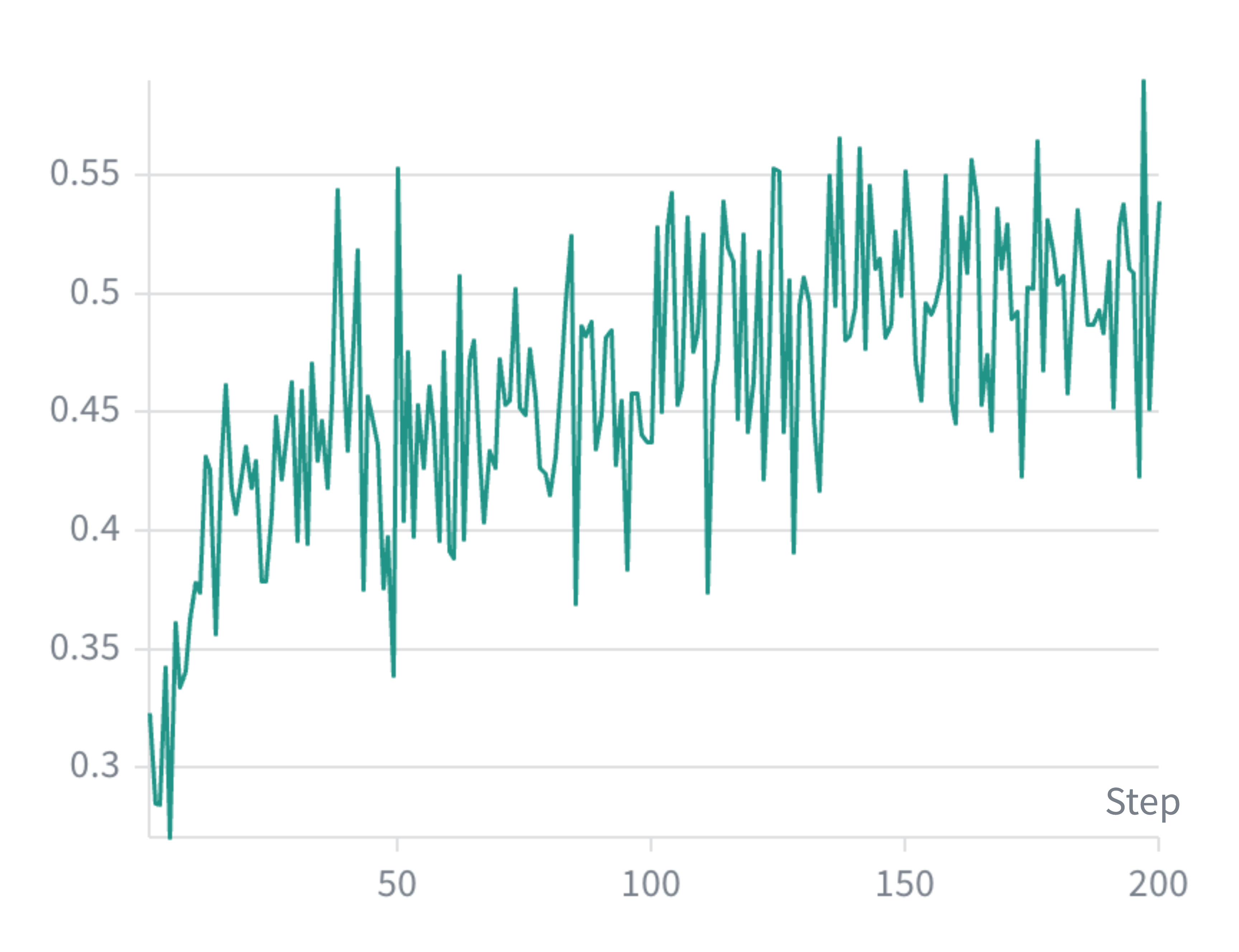} \\[-2pt]
        \raisebox{0.4\imgheight}{\rotatebox{90}{\small Gemini}} &
        \includegraphics[width=0.32\textwidth]{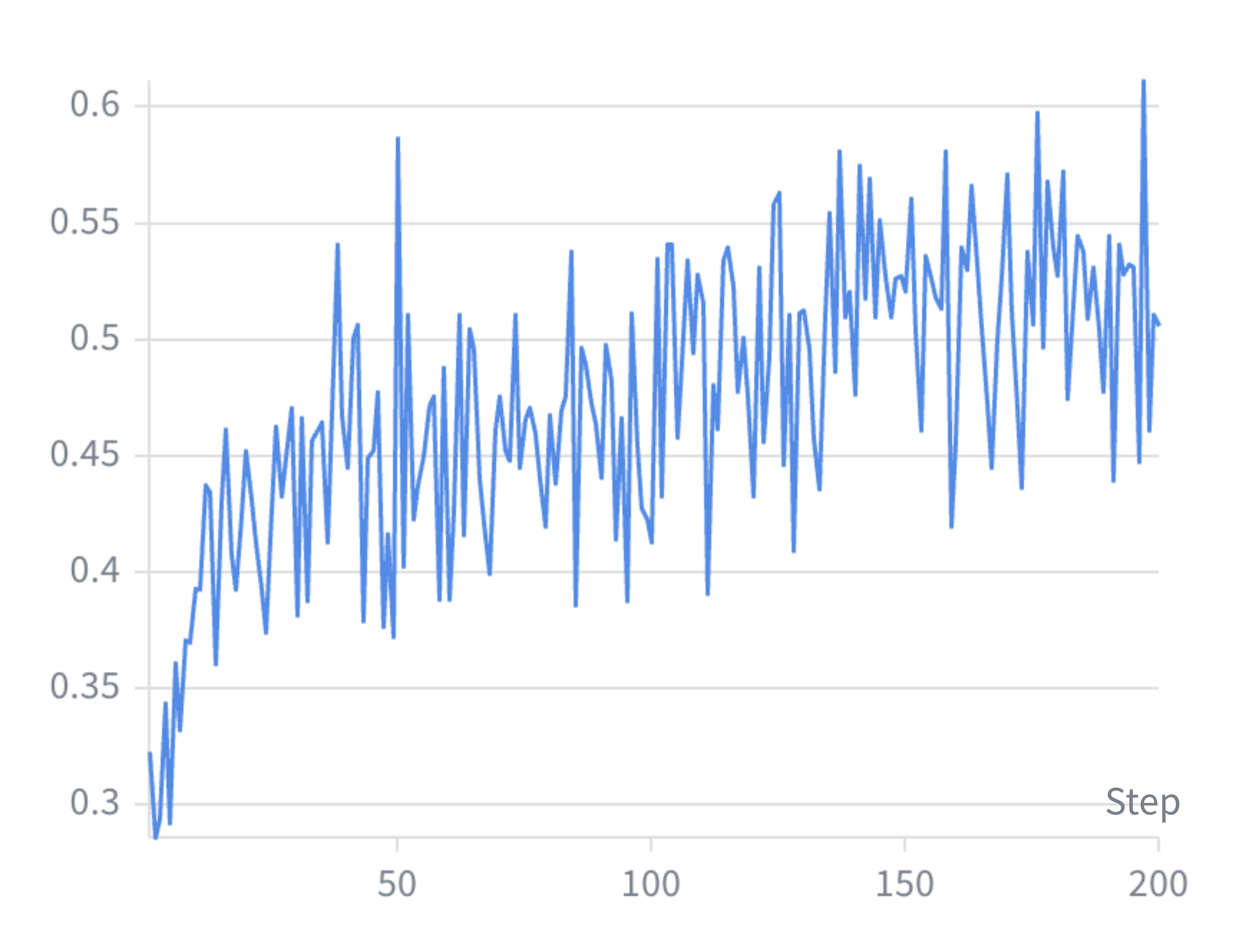} &
        \includegraphics[width=0.32\textwidth]{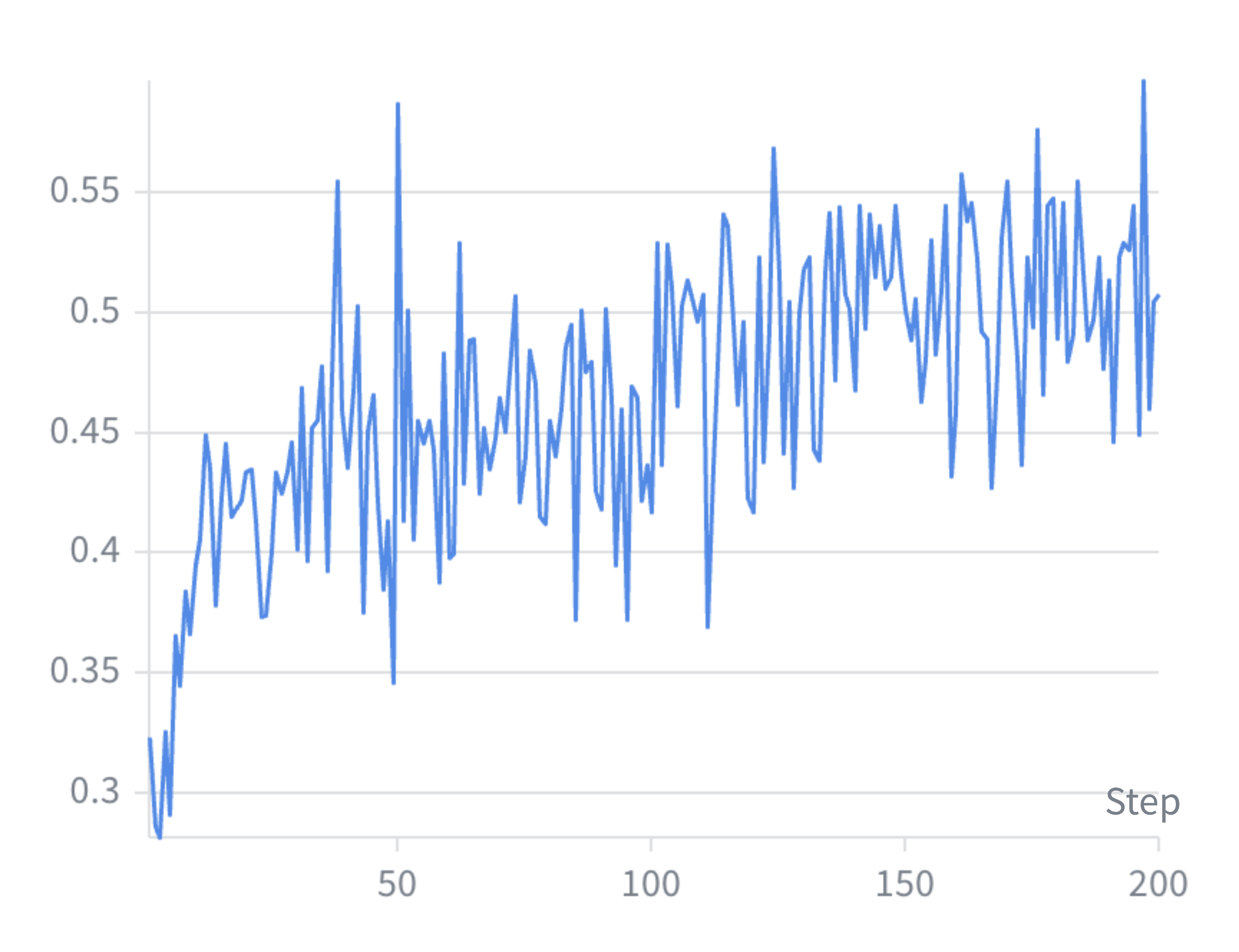} &
        \includegraphics[width=0.32\textwidth]{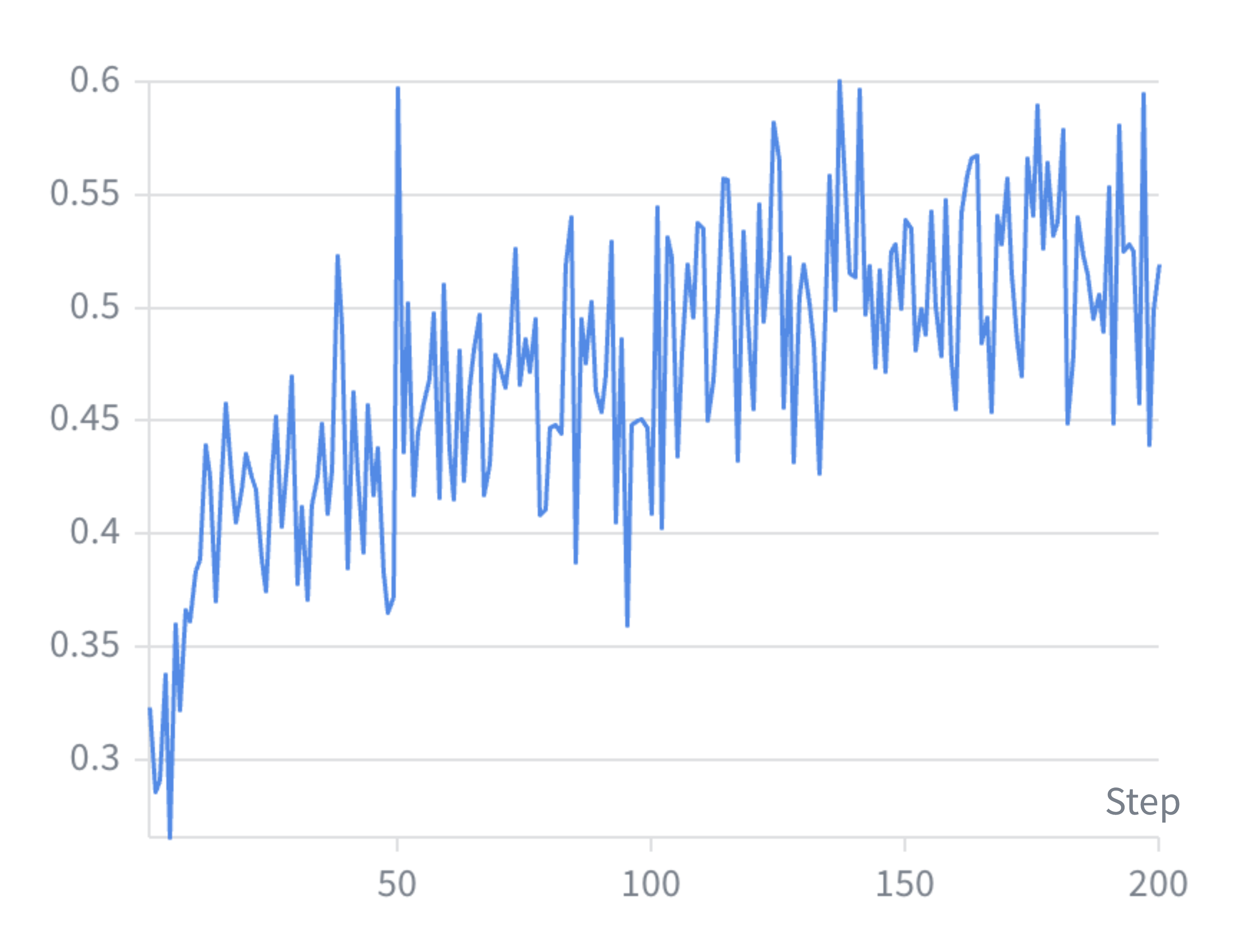} \\
      \end{tabular}
      \caption{\textbf{Training success rate} of \textbf{Qwen3-4B} under varying distillation weight $\lambda_{\text{OPSD}}$ with different proprietary teacher models.}
      \label{fig:training_4b}
  \end{figure}

% We present the training dynamics of MAPD across all model scales and distillation weight $\lambda_{OPSD}$ in Figures~\ref{fig:training_1.7b}, ~\ref{fig:training_4b}.
To provide a comprehensive overview of the optimization process, Figures~\ref{fig:training_1.7b} and \ref{fig:training_4b} illustrate the detailed training dynamics of MAPD. These visualizations track the learning trajectories across different model scales, while systematically highlighting how varying the distillation weight, $\lambda_{OPSD}$, influences the overall convergence and performance evolution\citep{jiang2026large, jia2026benchmarking, jiang2025mmke, qi2025context, Fu2026CanML, Fu2026MMKUBenchAM}.

\end{document}